\documentclass[11pt]{article}

\usepackage[preprint]{acl}

\usepackage{times}
\usepackage{latexsym}
\usepackage[T1]{fontenc}
\usepackage[utf8]{inputenc}
\usepackage{microtype}
\usepackage{inconsolata}

\usepackage{amsmath}
\usepackage{amsfonts}
\usepackage{amssymb}
\usepackage{booktabs}
\usepackage{array}
\usepackage{multirow}
\usepackage{tabularx}
\usepackage{nicefrac}

\usepackage{graphicx}
\usepackage{subcaption}
\usepackage{float}
\usepackage{placeins}

\usepackage[table]{xcolor}
\usepackage[most]{tcolorbox}
\usepackage{colortbl}

\usepackage{url}
\usepackage{hyperref}
\usepackage{ragged2e}

\usepackage{CJKutf8}
\newcommand{\zh}[1]{\begin{CJK*}{UTF8}{gbsn}#1\end{CJK*}}

\definecolor{casegreen}{RGB}{247,251,248}
\definecolor{caseblue}{RGB}{245,248,252}
\definecolor{casegreenline}{RGB}{203,222,210}
\definecolor{caseblueline}{RGB}{204,216,232}
\definecolor{okgreen}{RGB}{44,138,85}
\definecolor{badred}{RGB}{181,68,68}
\definecolor{hl}{RGB}{255,241,168}

\definecolor{hlgreen}{RGB}{220,237,220}
\definecolor{hlred}{RGB}{250,228,228}
\definecolor{hlblue}{RGB}{225,235,250}
\definecolor{accentblue}{RGB}{45,85,150}
\definecolor{accentred}{RGB}{190,65,65}
\definecolor{accentgreen}{RGB}{60,130,70}
\definecolor{mutedgray}{RGB}{110,110,110}

\newcommand{\best}[1]{\textbf{\textcolor{accentblue}{#1}}}
\newcommand{\worst}[1]{\textcolor{accentred}{#1}}

\newtcbox{\hltext}{on line, boxrule=0pt, boxsep=1pt,
  left=1pt, right=1pt, top=1pt, bottom=1pt,
  colback=hl, colframe=hl, arc=1pt}

\newtcolorbox{casebox}[2][]{
  enhanced,
  colback=#1,
  colframe=#2,
  boxrule=0.7pt,
  arc=2mm,
  left=1.8mm,
  right=1.8mm,
  top=1.2mm,
  bottom=1.2mm,
  before skip=0.45em,
  after skip=0.45em,
  fontupper=\RaggedRight\footnotesize
}

\title{Disentangling Language Roles in Multilingual LLM Task Execution}

\author{
Qishi Zhan\thanks{Corresponding author: \texttt{qishi.zhan@marquette.edu}.} \\
Marquette
\And
Minxuan Hu \\
Cornell
\And
Seoyeon Jang \\
UC San Diego
\AND
Lei Zhao \\
UPenn
\And
Ziheng Chen \\
UT Austin
\And
Man Liang \\
Maryland
\AND
Xinyue Xiang \\
Michigan
\And
Jiaxin Liu \\
UIUC
\And
Guansu Wang \\
Melbourne
\And
Liang He \\
Stanford
}

\begin{document}

\maketitle

\begin{abstract}
Multilingual LLMs are increasingly used when instruction, source content, and
required response languages do not coincide. Existing benchmarks have expanded
multilingual instruction-following evaluation, but they rarely isolate these
three roles within a fully crossed design. We introduce MTM-Bench, a controlled
benchmark for language-conditioned task execution in which each instance is
defined by a triplet
\((L_{\text{instr}}, L_{\text{content}}, L_{\text{resp}})\). Across English,
Spanish, and Chinese, MTM-Bench enumerates all 27 triplets and contains
2{,}430 instances per model across semantic reversal, final-state extraction,
and language purity with update realization. We evaluate 20 frontier and
open-weight LLMs using decomposed metrics for semantic correctness,
target-language adherence, constraint satisfaction, contamination ratio, and
joint success, with scoring validated by a targeted human audit. The fully
crossed design reveals that degradation is organized by the role a language
occupies in the task structure, not merely by mismatch count. The
response-language role is the dominant axis of variation, and a single
response-slot mismatch accounts for most degradation. The response-only and
full-mismatch comparison suggests that mismatch count is not a monotonic
predictor of difficulty, with model-level ordering varying across systems. Task
families fail through distinct channels, showing that semantic correctness
alone does not capture reliable multilingual task execution.
\end{abstract}

\section{Introduction}

\begin{figure*}[t]
  \centering
  \includegraphics[width=\textwidth]{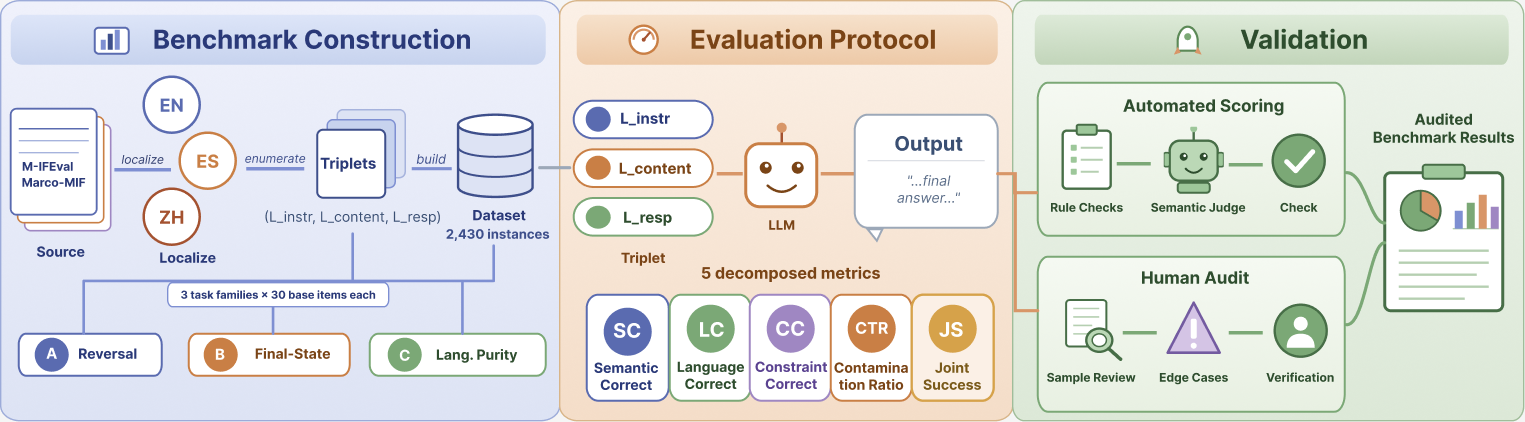}
  \caption{Overview of MTM-Bench: benchmark construction, decomposed evaluation,
  and scoring validation.}
  \label{fig:overview}
\end{figure*}

Large language models are increasingly used in multilingual workflows where
language roles are no longer aligned. A Spanish-speaking employee in a
multinational company may instruct an LLM in Spanish, provide an English email
thread as source content, and ask for a Chinese summary to share with colleagues
in China. Such cases make multilingual task execution a triplet-structured
problem: the instruction language, content language, and response language may
each differ. Successful execution therefore requires more than general
multilingual fluency. A model must understand the task from the instruction,
process the source content, realize the answer in the specified response
language, and follow explicit output constraints.

Despite its practical importance, this setting remains difficult to evaluate
cleanly. Existing multilingual benchmarks have expanded language coverage,
cross-lingual transfer evaluation, and instruction-following assessment.
However, they often do not treat the instruction language, source content
language, and required response language as independently controlled roles
within a single evaluation design
~\citep{conneau2018xnli,lewis2020mlqa,hu2020xtreme,ruder2021xtremer,liang2020xglue,fitzgerald2023massive,ahuja-etal-2023-mega,bandarkar-etal-2024-belebele}.
As a result, when a model fails, it can be unclear whether the error arises from
task misunderstanding, content processing, target-language realization,
constraint following, or an interaction among these factors.

% We investigate this problem through language-conditioned task execution, treating each instance as an explicit triplet \((L_{\text{instr}}, L_{\text{content}}, L_{\text{resp}})\) that specifies the instruction language, content language, and required response language independently.

We investigate this problem through language-conditioned task execution. Each
instance is defined by an explicit triplet
\((L_{\text{instr}}, L_{\text{content}}, L_{\text{resp}})\), representing the
instruction language, content language, and required response language. 
This formulation treats multilingual execution as a structured interaction among three language roles, separating task understanding from output realization.

We introduce MTM-Bench (Multilingual Triplet Mismatch Benchmark), a controlled
benchmark for this setting (Figure~\ref{fig:overview}). Across English
(EN), Spanish (ES), and Chinese (ZH), we enumerate all \(3^3=27\) language
triplets and construct 2{,}430 instances per model over three task families:
semantic reversal, final-state extraction, and language purity with update
realization. We evaluate model outputs with decomposed metrics for semantic
correctness, target-language adherence, explicit constraint satisfaction,
contamination ratio, and joint success. 

MTM-Bench makes three contributions. First, it reframes multilingual task
execution as a triplet-based evaluation problem, where the instruction, content,
and response languages are varied independently rather than treated as one
undifferentiated language variable. Second, it offers a fully crossed EN/ES/ZH
benchmark spanning 27 language triplets and three task families, scored with a
decomposed protocol that separates semantic correctness, target-language
adherence, constraint satisfaction, contamination, and joint success, and
checked against a targeted human audit. Third, the fully crossed design enables role-specific marginal comparisons that
aggregate multilingual benchmarks cannot isolate: response-language variation is
much larger than instruction- or content-language variation, and the
response-only versus full-mismatch contrast suggests mismatch count is not a
monotonic predictor of difficulty.

\section{Related Work}

Cross-linguistic instruction following has become a central testbed for
multilingual LLMs~\citep{conneau2020xlmr,xue2021mt5,wei2022finetuned,ouyang2022training}.
Prior work shows that instruction-following behavior can transfer across
languages, but transfer varies by language, model, and tuning mixture
~\citep{ranaldi2024crossalpaca,muennighoff2023crosslingual,shaham2024multilingual}.
Our focus is diagnostic rather than training-oriented: we ask where multilingual
task execution fails when instruction, content, and response languages vary
independently.

Recent instruction-following benchmarks have expanded both language coverage and
evaluation realism. IFEval introduced automatically verifiable
instruction-following evaluation in English~\citep{zhou2023ifeval}, while
FollowBench examines fine-grained constraint following across content,
situation, style, format, and example dimensions~\citep{jiang2024followbench}.
Subsequent work has extended this agenda in multiple directions: M-IFEval
adapts the framework to multilingual settings and documents substantial
language-dependent variation~\citep{dussolle2025mifeval}; MaXIFE scales coverage
to many languages and verifiable tasks~\citep{liu2025maxife}; and
Marco-Bench-MIF emphasizes verifiable execution under language-specific
constraints~\citep{zeng2025marco}. XIFBench evaluates instruction following
through constraint categories and requirement-based validation, while Multi-IF
moves the setting into multi-turn
conversations~\citep{li2025xifbench,he2024multiif}.

These benchmarks provide strong task templates, language-specific constraints,
and verifiable evaluation protocols. MTM-Bench addresses a complementary
diagnostic question. Prior work asks how well a model follows instructions in each language overall;
MTM-Bench asks which language role drives failure when the instruction language,
source content language, and required response language vary independently. This role-based design separates the target
language itself from mismatch between earlier prompt context and the required
answer language, and from the number of languages present in an instance.
MTM-Bench therefore retains the emphasis on verifiable tasks while reorganizing
evaluation around role-specific language variation. This fully crossed design
enables role-specific marginal comparisons that prior multilingual
instruction-following benchmarks cannot isolate: the same target task can be
evaluated while varying only the instruction, content, or response language
role.

Another related line of work addresses how to evaluate multilingual outputs.
CoCo-CoLa argues that concept-level correctness and target-language correctness
should be evaluated separately rather than collapsed into a single decision
~\citep{rahmati2025cococola}. Language-control failures have also been studied
as language confusion: ~\citet{marchisio2024languageconfusion} show that LLMs may fail to generate
outputs in the intended language even when that language is explicitly specified. These findings motivate our decomposed
evaluation protocol, which separates semantic correctness, target-language
adherence, constraint satisfaction, and non-target-language contamination. 

\section{Framework}
\label{sec:framework}

An evaluation instance is defined as
\[
x=(i,c,y^\star,L_{\mathrm{instr}},L_{\mathrm{content}},L_{\mathrm{resp}}),
\]
where \(i\) is the instruction, \(c\) is the source content, \(y^\star\) is the
reference answer or target condition, and
\((L_{\mathrm{instr}},L_{\mathrm{content}},L_{\mathrm{resp}})\) specifies the
instruction language, content language, and required response language. Given
\(x\), a model produces an output \(\hat y\).

We evaluate each output along three binary dimensions:
\[
\begin{aligned}
\textsc{SC} &= \textsc{SemanticCorrect}(\hat y,y^\star),\\
\textsc{LC} &= \textsc{LangCorrect}(\hat y,L_{\mathrm{resp}}),\\
\textsc{CC} &= \textsc{ConstraintCorrect}(\hat y).
\end{aligned}
\]
\textsc{SC} measures whether the output captures the target meaning,
\textsc{LC} measures whether it is realized in the required response language,
and \textsc{CC} measures whether it satisfies explicit output constraints.

We also define \textsc{ContaminationRatio} (\textsc{CTR}) as a continuous
measure of non-target-language material. Let \(\mathcal{A}\) be an allowlist of
language-neutral units, including pure numerals, single-letter labels,
allowlisted proper names, and allowlisted platform or system names. We define scored units as lexical spans, including contiguous Latin alphabetic
tokens, contiguous Chinese-character spans, and numeric spans after applying the
allowlist. Let \(N_{\mathrm{nt}}(\hat y,L_{\mathrm{resp}})\) be
the number of non-target scored units after excluding \(\mathcal{A}\), and let
\(N_{\mathrm{all}}(\hat y)\) be the number of scored units. We compute
\[
\textsc{CTR}(\hat y,L_{\mathrm{resp}})
=
\frac{N_{\mathrm{nt}}(\hat y,L_{\mathrm{resp}})}
     {N_{\mathrm{all}}(\hat y)}.
\]

\textsc{LC} and \textsc{CTR} capture complementary aspects of language control.
\textsc{LC} is a binary admissibility judgment: it records whether the response
satisfies the target-language requirement under the allowlist. \textsc{CTR}
provides a graded severity measure, distinguishing isolated carryover from
substantial non-target-language contamination.
The aggregate success metric is
\[
\textsc{JS}=\textsc{SC}\cdot\textsc{LC}\cdot\textsc{CC}.
\]
The multiplicative definition treats the three requirements as jointly necessary
rather than compensatory: a response is not fully successful if it solves the
task but uses the wrong response language, or if it is in the target language
but violates the requested output form. Operational scoring rules are described
in \S~\ref{sec:evaluation}.

\section{Benchmark}

We construct MTM-Bench to support controlled comparisons across instruction,
content, and response language roles while keeping task semantics fixed. The benchmark is designed around four constraints: semantic portability
across EN, ES, and ZH; short outputs that make language-control failures
observable; closed-form or rubric-verifiable answers; and separability of
semantic correctness, target-language correctness, and format compliance. These
constraints motivate our focus on short, controlled tasks rather than
open-ended multilingual generation
~\citep{ribeiro-etal-2020-beyond,naik-etal-2018-stress}.

We use EN, ES, and ZH as a compact multilingual testbed for the
fully crossed triplet design. EN  serves as a high-resource baseline
language in many existing LLM evaluations, ES retains the Latin
script while differing from English in morphology, function words, and surface
realization, and ZH introduces a non-Latin script
and substantial typological distance from EN and ES~\citep{pires-etal-2019-multilingual}. This combination provides complementary contrasts while keeping
benchmark construction, localization, and scoring validation tightly controlled.
We next describe how the benchmark is constructed under this design.

\subsection{Benchmark Construction}
\label{sec:benchmark_construction}

We construct MTM-Bench through a controlled rewriting and localization pipeline.
We begin from task patterns in prior multilingual instruction-following
benchmarks, especially M-IFEval and Marco-Bench-MIF~\citep{dussolle2025mifeval,zeng2025marco}, and retain patterns that
are semantically portable across EN, ES, and ZH and admit closed-form or
rubric-verifiable evaluation. For semantic-reversal items, prior irony and reversal resources guide the
phenomenon categories and item patterns. We rewrite each item as a short controlled scenario in a triplet-based format
with normalized metadata, gold conditions, and language-specific output
constraints, rather than directly translating existing prompts, so that the
intended stance remains stable across all triplet variants.

Each selected pattern is instantiated as a base item with separable instruction,
content, and response-language components. This design allows the same semantic
target to be evaluated under any
\((L_{\text{instr}}, L_{\text{content}}, L_{\text{resp}})\) triplet. Across
triplet variants, names, dates, times, room numbers, platform names, and answer
formats are held fixed when they define the underlying task, while ordinary
lexical and functional material is localized according to the relevant language
slot.

We manually check that each base item has a unique gold answer or target
condition, that distractor values are not accidentally valid, and that changing
one language slot does not alter the semantic target. Ambiguous or unstable
items are removed or rewritten. The final benchmark contains 30 base items per
task family. For each base item, we instantiate all 27 combinations of
instruction, content, and response languages, yielding 810 instances per family
and 2{,}430 instances per model.

\subsection{Task Families}
\label{sec:task_families}

We organize the benchmark into three task families, each targeting a distinct
multilingual execution challenge. The families are chosen to cover three
complementary points in the execution pipeline: recovering intended meaning
despite misleading surface form, extracting the final valid state under explicit
format constraints, and realizing an updated state in the requested response
language. 
% This choice lets us test semantic resolution, closed-form extraction,
% and target-language realization within the same triplet framework. 
Family A focuses on semantic reversal. Each item presents surface wording that suggests one interpretation, while the
author's intended stance points in the opposite direction or substantially
revises it. This family includes irony, sarcasm, understatement, ostensible
praise masking criticism, and concluding remarks that overturn earlier
framing~\citep{vanhee2018irony}. The task tests whether a model can identify
the underlying polarity and evaluative target rather than relying on surface
wording alone.

Family B focuses on final-state extraction. Each item presents a short
administrative or informational text containing multiple candidate values, such
as times, deadlines, locations, contacts, or requirements. Only the final valid
value is correct after updates, corrections, rescheduling, or reassignment. This
family follows the verifiable, closed-form extraction style emphasized in prior
multilingual instruction-following benchmarks~\citep{dussolle2025mifeval,liu2025maxife,zeng2025marco}, but is rewritten into our
triplet-based setting and scored with a family-specific final-value rubric.

Family C focuses on language purity with update realization in short
administrative scenarios. Each item presents a controlled update, such as a
meeting room change, reassigned contact person, revised platform, or moved
deadline, and asks the model to describe the resulting update in the required
response language. By language purity, we mean that ordinary lexical and
functional material must be realized in \(L_{\text{resp}}\). Proper nouns,
numeric expressions, and allowlisted platform or system names, such as Canvas,
Zoom, Teams, and Moodle, may remain unchanged. 
Non-target language framing must be localized: in a ZH response, for example,
``Room 327'' and ``Friday at 4 PM'' are treated as EN structural carryover,
whereas ``Zoom\zh{会议改到周五下午4点}''
(`the Zoom meeting is moved to Friday at 4 PM')
is allowed because the platform name is allowlisted and the surrounding
structure is localized.
% Non-target language framing must
% be localized: in a ZH response, for example, ``Room 327'' and ``Friday at 4 PM''
% are treated as EN structural carryover, whereas ``Zoom\zh{会议改到周五下午4点}''
% is allowed because the platform name is allowlisted and the surrounding
% structure is localized. 
We treat non-localized structural carryover as a failure
because the task explicitly asks for the response to be localized into
\(L_{\text{resp}}\); the criterion is not meant to judge the appropriateness of
code-switching when users permit or prefer mixed-language
use~\citep{dogruoz2021survey}. See Appx.~\ref{app:rubric} for the complete
rubric for each task family.

\section{Evaluation Protocol}
\label{sec:evaluation}

We score each output using the decomposed metrics defined in
\S~\ref{sec:framework}: \textsc{SemanticCorrect}, \textsc{LangCorrect},
\textsc{ConstraintCorrect}, \textsc{ContaminationRatio}, and
\textsc{JointSuccess}.

\textsc{SemanticCorrect} is evaluated by task family. For Family A, the output
must recover the writer's intended stance rather than the surface wording. For
Family B, only the final valid value receives credit, with format-equivalent or
cross-linguistic variants accepted when they refer to the same final value. For
Family C, the output is correct if it captures the main update without
contradiction. Rule-based prefilters handle obvious failures, including empty outputs and
explicit non-answers, as well as family-specific failures such as polarity
mismatches or missing critical numeric anchors. We then score Family B with a deterministic final-value checker, while
Families A and C use an LLM-assisted judge with family-specific rubrics
~\citep{kocmi-federmann-2023-large,liu2023geval,zheng2023judging}. The judge
receives the benchmark item, the model output, and the task-specific semantic
rubric, but not the identity of the evaluated model. Language correctness and
constraint correctness are evaluated separately from semantic correctness, so
the judge is not used to determine target-language adherence or explicit
format compliance.

\textsc{LangCorrect} applies a strict target-language criterion. Pure numerals,
single-letter labels, allowlisted proper names, and allowlisted platform names
are treated as language-neutral. Non-target lexical items or structural frames
cause failure unless they fall under this allowlist.
% For example, ``327'' is language-neutral in ZH, but ``Room 327'' fails because
% \textit{Room} is an English structural token. Representative edge cases are
% reported in Appx.~\ref{app:edge}.
For instance, ``327'' passes as language-neutral in a ZH-target response,
whereas ``Room 327'' fails because the English structural token
\textit{Room} is not allowlisted. Representative edge-case decisions are reported in Appx.~\ref{app:edge}.

\textsc{ConstraintCorrect} checks explicit output requirements in the
instruction. We implement rule-based checks for five recurring constraint
types: yes-or-no with brief reason, output-only, single sentence, banned
word, and starts-with prefix. For output-only tasks, explanatory prefixes
and continuation-style explanations are marked as violations, while minor
formatting variation is allowed when it does not change the final value.

\textsc{ContaminationRatio} operationalizes the framework-level definition by
counting non-target scored units after applying the allowlist. For ZH targets,
non-allowlisted Latin alphabetic tokens are treated as non-target units. For ES
targets, non-allowlisted EN structural tokens, venue words, month names, and
common function words are treated as non-target units. This metric is related to
prior work on code-switching and language confusion, but is tailored to short,
constraint-driven outputs rather than open-ended code-switched text
~\citep{aguilar2020lince,dogruoz2021survey,marchisio2024languageconfusion}.

We validate the scoring pipeline through a human audit covering task families,
response languages, triplet configurations, and major failure types. The audit
contains two complementary subsets: 1022 targeted boundary cases marked during
scoring development and validation, and 300 unflagged routine outputs sampled
as a sanity check. 

Human annotators inspect the benchmark item, model response,
and automatic component labels, and record overrides for \textsc{SemanticCorrect}, \textsc{LangCorrect}, and
\textsc{ConstraintCorrect}. The targeted set has a component-level override rate
of 18.2\%, while the unflagged check has a lower override rate of 7.3\%,
suggesting that most routine cases are handled consistently by the automatic
pipeline. At the \textsc{JointSuccess} level, the targeted-set override rate is
lower, at 15.4\%, because some component corrections do not alter the final
joint decision. Disagreements are adjudicated using the edge-case policy in
Appx.~\ref{app:edge}~\citep{artstein2008inter}; full audit details, including
per-language agreement rates, are reported in Appx.~\ref{app:human}.

\section{Experimental Setup}

We evaluate 20 models spanning frontier commercial systems, open-weight models, and different degrees of multilingual specialization: Claude Haiku 4.5, Claude Sonnet 4.6, and Claude Opus 4.1 \citep{anthropic2025claudehaiku45,anthropic2026claudesonnet46,anthropic2025claudeopus41}; GPT-4.1, GPT-5.4, GPT-5.4-mini, and GPT-5.3-Codex \citep{openai2025gpt41,openai2026gpt54,openai2026gpt54mini,openai2026gpt53codex}; Gemini 2.5 Flash and Gemini 2.5 Flash-Lite \citep{comanici2025gemini25,googledeepmind2025gemini25flashlite}; DeepSeek-V3.1, DeepSeek-V3.2, and DeepSeek-R1 \citep{deepseekai2025deepseekv31,deepseekai2025deepseekv32,deepseekai2025deepseekr1}; Qwen3-Max, Qwen3.5-Plus, and Qwen3.5-Flash \citep{yang2025qwen3technicalreport,alibabacloud2026modelstudioqwen}; MiniMax-M2.5 and MiniMax-M2.7 \citep{minimax2026m25,minimax2026m27}; and LLaMA-3, LLaMA-3.1-8B, and LLaMA-3.3-70B \citep{grattafiori2024llama3,meta2024llama33modelcard}.

% We evaluate 20 models spanning frontier commercial systems, open-weight models,
% and different degrees of multilingual specialization: Claude Haiku 4.5, Claude
% Sonnet 4.6, Claude Opus 4.1; GPT-4.1, GPT-5.4, GPT-5.4-mini, GPT-5.3-Codex;
% Gemini 2.5 Flash, Gemini 2.5 Flash-Lite; DeepSeek-V3.1, DeepSeek-V3.2,
% DeepSeek-R1; Qwen3-Max, Qwen3.5-Plus, Qwen3.5-Flash; MiniMax-M2.5,
% MiniMax-M2.7; and LLaMA-3, LLaMA-3.1-8B, LLaMA-3.3-70B.

We evaluate all models on the full 2{,}430-instance benchmark
(3 task families $\times$ 30 base items $\times$ 27 triplets) in a zero-shot
setting, without system prompts or few-shot demonstrations~\citep{brown2020language,wei2022finetuned}. For each instance,
we use a fixed prompt template that concatenates the localized instruction,
source content, and question, and we request one response per model. We set
temperature to 0 when supported and leave top-$p$ at the provider default~\citep{holtzman2020curious}. We do
not enable optional reasoning or thinking modes unless they are part of the
model's default public interface. Full generation details are reported in
Appx.~\ref{app:generation}. 

\begin{table*}[t]
\centering
\small
\renewcommand{\arraystretch}{1.1}
\setlength{\tabcolsep}{12pt}
\begin{tabular}{lrrrrr}
\toprule
\textbf{Model} & \textbf{SC}~$\uparrow$ & \textbf{LC}~$\uparrow$ & \textbf{CC}~$\uparrow$ & \textbf{CTR}~$\downarrow$ & \textbf{JS}~$\uparrow$ \\
\midrule
\rowcolor{hlgreen}
Qwen3-Max             & \best{.937} & \best{.949} & \best{.925} & .021        & \best{.840} \\
\rowcolor{hlgreen}
DeepSeek-R1           & .928        & .946        & .923        & \best{.019} & \best{.822} \\
\rowcolor{hlgreen}
Qwen3.5-Plus          & .913        & \best{.958} & \best{.932} & .021        & \best{.820} \\
GPT-5.4-mini          & \best{.935} & .935        & .908        & .023        & .815 \\
GPT-5.4               & .930        & \best{.949} & .893        & \best{.016} & .799 \\
DeepSeek-V3.2         & .909        & .945        & .905        & .024        & .796 \\
DeepSeek-V3.1         & .908        & .927        & .901        & .037        & .786 \\
GPT-5.3-Codex         & \best{.938} & .944        & .869        & \best{.020} & .781 \\
Claude Sonnet 4.6     & .926        & .930        & .886        & .033        & .767 \\
Gemini 2.5 Flash-Lite & .872        & .944        & \best{.924} & .023        & .767 \\
Qwen3.5-Flash         & .911        & .878        & .908        & .068        & .765 \\
GPT-4.1               & .912        & .911        & .886        & .045        & .758 \\
Claude Opus 4.1       & .918        & .910        & .876        & .043        & .741 \\
LLaMA-3.3-70B         & .880        & .893        & .915        & .068        & .740 \\
Gemini 2.5 Flash      & .867        & .919        & .899        & .043        & .727 \\
MiniMax-M2.7          & .851        & .847        & .820        & .034        & .719 \\
MiniMax-M2.5          & .800        & .796        & \worst{.786} & .041       & .681 \\
\rowcolor{hlred}
Claude Haiku 4.5      & .925        & .906        & \worst{.795} & .051       & \worst{.678} \\
\rowcolor{hlred}
LLaMA-3               & \worst{.830} & \worst{.790} & .799       & \worst{.131} & \worst{.573} \\
\rowcolor{hlred}
LLaMA-3.1-8B          & \worst{.761} & \worst{.817} & .839       & \worst{.105} & \worst{.537} \\
\midrule
\textit{Mean}         & .893 & .905 & .880 & .043 & .745 \\
\bottomrule
\end{tabular}
\caption{Main results for 20 models. Top-3 per column in
\textbf{\textcolor{accentblue}{blue}}, bottom-3 in \textcolor{accentred}{red}.
Sorted by \textsc{JointSuccess}.}
\label{tab:main_results}
\end{table*}

\section{Results}
\label{sec:results}

\subsection{Overall Performance}

Across the 20 models, \textsc{JointSuccess} ranges from 0.537 to 0.840
(mean 0.745; Table~\ref{tab:main_results}). The component metrics are
higher in isolation: \textsc{LangCorrect} has mean 0.905,
\textsc{SemanticCorrect} has mean 0.893, and \textsc{ConstraintCorrect} has mean
0.880. Satisfying all three simultaneously is substantially harder, showing that
multilingual task success depends on the joint satisfaction of semantic,
language, and constraint requirements.

Qwen3-Max achieves the highest overall \textsc{JointSuccess} (0.840), followed
by DeepSeek-R1 (0.822) and Qwen3.5-Plus (0.820). The overall ranking does not
reduce to any single component metric. GPT-5.3-Codex achieves the highest
\textsc{SemanticCorrect} across all 20 models yet ranks lower on
\textsc{ConstraintCorrect}; Gemini 2.5 Flash-Lite shows the opposite profile.
These cases illustrate why decomposed evaluation is needed: a model can recover
the correct meaning while consistently failing format requirements, or satisfy
output-form constraints while making more semantic
errors~\citep{liang2023holistic}.

\begin{table}[!t]
\centering
\small
\setlength{\tabcolsep}{3.6pt}
\renewcommand{\arraystretch}{1.06}
\begin{tabular}{@{}lrrrrrr@{}}
\toprule
\textbf{Class} & \textbf{n} & \textbf{SC} & \textbf{LC}
  & \textbf{CC} & \textbf{CTR} & \textbf{JS} \\
\midrule
\rowcolor{hlblue}
Same          & 270 & \best{.920} & \best{.975} & \best{.910} & \best{.008} & \best{.829} \\
Instr only    & 540 & .909 & .939 & .888 & .022 & .795 \\
Content only  & 540 & .890 & .913 & .891 & .039 & .746 \\
Resp only     & 540 & .880 & \worst{.860} & \worst{.860} & \worst{.065} & .694 \\
Full mismatch & 540 & .878 & .871 & .864 & .064 & .705 \\
\midrule
\multicolumn{7}{@{}l}{\textit{By task family}} \\
\rowcolor{hlblue}
A: Sem. rev.    & 810 & .925 & .888 & .906 & \worst{.066} & \best{.786} \\
B: Final-state  & 810 & \best{.937} & \best{.948} & \worst{.796} & .036 & .728 \\
\rowcolor{hlred}
C: Lang. purity & 810 & \worst{.816} & .878 & \best{.936} & \best{.027} & \worst{.722} \\
\bottomrule
\end{tabular}
\caption{Performance by mismatch class and task family, averaged across all
20 models. Blue shading marks the highest \textsc{JointSuccess} row in each
block; red shading marks the lowest task-family row. Response-only and
full-mismatch triplets are compared in \S~\ref{sec:role_position}.}
\label{tab:mm_class}
\end{table}

Table~\ref{tab:mm_class} summarizes performance by mismatch class and task
family, providing the aggregate views used below to separate language-role
effects from task-family effects.

\subsection{Response Language Is the Dominant Axis}

\begin{figure*}[t]
  \centering
  \includegraphics[width=\linewidth]{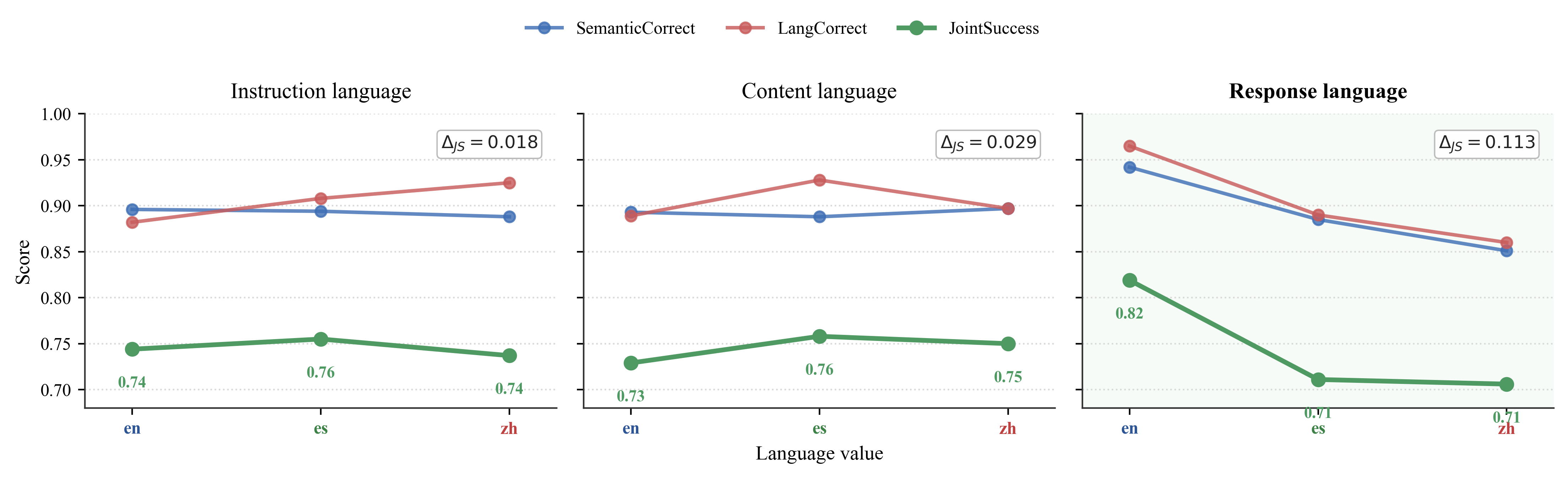}
  \caption{Marginal effects of the three language roles on
\textsc{SemanticCorrect}, \textsc{LangCorrect}, and \textsc{JointSuccess}
across EN, ES, and ZH. The response axis has the largest
\textsc{JointSuccess} range (\(\Delta_{\mathrm{JS}}=0.113\)).}
  \label{fig:axes}
\end{figure*}

The response language is the dominant axis of variation. As shown in
Table~\ref{tab:axes} and Figure~\ref{fig:axes}, \textsc{JointSuccess} in the
response slot drops from 0.819 for EN to 0.711 for ES and 0.706 for ZH, a range
of \(\Delta_{\mathrm{resp}}=0.113\). The corresponding ranges are much smaller
for content language (\(\Delta_{\mathrm{content}}=0.029\)) and instruction
language (\(\Delta_{\mathrm{instr}}=0.018\)). This ordering remains stable under clustered bootstrap resampling over base
items: the 95\% interval for the response-axis range is \([0.083, 0.156]\),
compared with \([0.012, 0.049]\) for the content axis and \([0.006, 0.033]\)
for the instruction axis. Bootstrap details are reported in
Appx.~\ref{app:stat_uncertainty}.

A fixed-effect model controlling for model identity and base-item difficulty
confirms the same ranking of the three axes. Response-slot coefficients are
substantially larger than instruction-slot and content-slot coefficients,
indicating that the response-role effect is not explained by model composition
or item difficulty. Content-slot effects are smaller and not directionally
consistent across ES and ZH, matching the marginal analysis in
Table~\ref{tab:axes}. Full regression results are reported in
Appx.~\ref{app:regression_analysis}.

\begin{table}[!t]
\centering
\small
\setlength{\tabcolsep}{4.2pt}
\renewcommand{\arraystretch}{1.03}
\begin{tabular}{@{}llrrrrr@{}}
\toprule
\textbf{Axis} & \textbf{Lang} & \textbf{SC} & \textbf{LC}
  & \textbf{CC} & \textbf{CTR} & \textbf{JS} \\
\midrule
\multirow{3}{*}{\shortstack[l]{Instr.\\\(\Delta_{\mathrm{JS}}=.018\)}}
  & en & .896 & .882 & .875 & .043 & .744 \\
  & es & .894 & .908 & .898 & .051 & .755 \\
  & zh & .888 & .925 & .866 & .036 & .737 \\
\midrule
\multirow{3}{*}{\shortstack[l]{Content\\\(\Delta_{\mathrm{JS}}=.029\)}}
  & en & .893 & .889 & .874 & .036 & .729 \\
  & es & .888 & .928 & .880 & .033 & .758 \\
  & zh & .897 & .897 & .884 & .060 & .750 \\
\midrule
\multirow{3}{*}{\shortstack[l]{\textbf{Resp.}\\\(\Delta_{\mathrm{JS}}=.113\)}}
  & en & .942 & .965 & .887 & .023 & \best{.819} \\
  & es & .885 & .890 & .873 & .039 & .711 \\
  & zh & .851 & .860 & .878 & .067 & \worst{.706} \\
\bottomrule
\end{tabular}
\caption{Marginal effects of language axes. Values are averaged over all
models and over the other two axes. \(\Delta\) denotes the range across
en/es/zh.}
\label{tab:axes}
\end{table}

The same ordering appears for \textsc{SemanticCorrect} and
\textsc{LangCorrect}: in Table~\ref{tab:axes}, the response axis has ranges of
0.091 and 0.105, respectively, while the other two axes are flatter.
\textsc{ContaminationRatio} moves in the opposite direction: ZH-response
instances have an average ratio of 0.067, compared with 0.023 for EN-response
instances (Table~\ref{tab:axes}). Model-level sensitivity patterns are reported
in Appx.~\ref{app:model_family}, Fig.~\ref{fig:model_role_sensitivity}. We next
ask whether aggregate degradation can be explained by mismatch count alone, or
whether the role in which a mismatch occurs provides additional explanatory
structure.

\subsection{Role Position Beyond Mismatch Count}
\label{sec:role_position}

To separate mismatch count from language-role position, we group the 27 triplets
into five classes: \textit{same}, \textit{instr-only}, \textit{content-only},
\textit{resp-only}, and \textit{full mismatch}. Table~\ref{tab:mm_class}
shows that performance drops from \textit{same} (0.829) to
\textit{instr-only} (0.795) and \textit{content-only} (0.746), but the lowest
scores occur when the response language is mismatched: \textit{resp-only}
reaches 0.694 and \textit{full mismatch} reaches 0.705.

If mismatch count alone were a monotonic predictor of task difficulty,
full-mismatch triplets would be expected to show a clear additional penalty
beyond the response-only single-mismatch case. A clustered bootstrap over base
items gives a full-minus-response-only difference of 0.011 with a 95\% interval
of [0.001, 0.022] (Appx.~\ref{app:stat_uncertainty}). Although this interval
lies slightly above zero, the difference is small and runs counter to the
direction expected under a monotonic mismatch-count account: full-mismatch
triplets do not score below response-only triplets. This suggests that mismatch
count is not a monotonic predictor of difficulty, and that much of the aggregate
degradation associated with multilingual mismatch is already present once the
required response language differs from the preceding language context.

This aggregate ordering is not consistent across models. We therefore do not
interpret the response-only and full-mismatch comparison as a directional claim
that full mismatch is intrinsically easier. Instead, we use it as evidence that
mismatch count alone does not account for the observed aggregate degradation
pattern. Model-level details are reported in
Appx.~\ref{app:mismatch_heterogeneity}.

\begin{figure}[!t]
  \centering
  \includegraphics[width=\columnwidth]{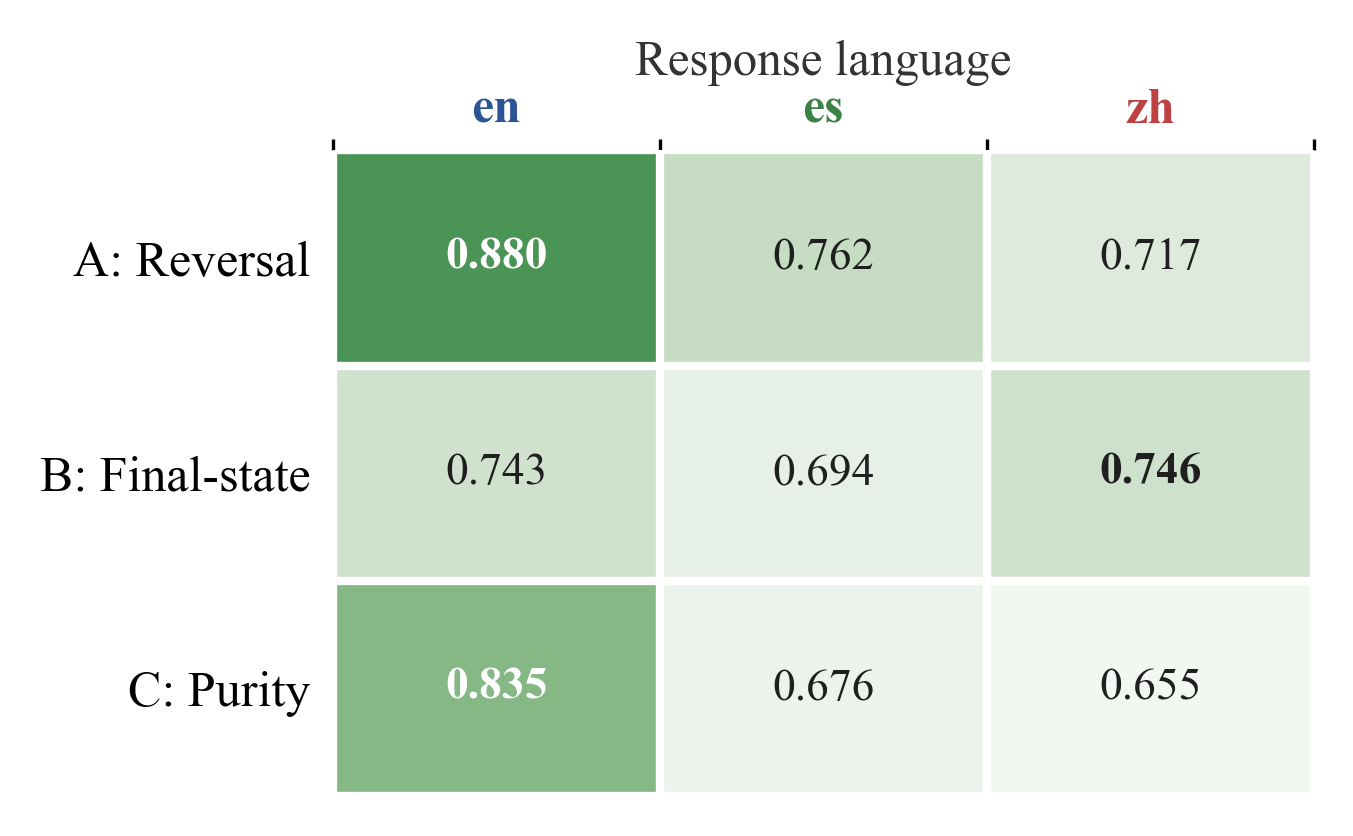}
  \caption{\textsc{JointSuccess} by task family and response language, averaged
  over 20 models. Final-state extraction is the main exception to the overall
  response-language degradation pattern.}
  \label{fig:cat_resp}
\end{figure}

\subsection{Task Families Expose Distinct Bottlenecks}

The three task families have similar overall \textsc{JointSuccess}
(0.722--0.786; Table~\ref{tab:mm_class}), but they fail through different
channels. Task~A has high \textsc{SemanticCorrect} (0.925), suggesting that
models often recover the intended stance, yet performance is limited by
response-language realization, especially for ZH responses
(Figure~\ref{fig:cat_resp}). Task~B has the highest \textsc{LangCorrect}
(0.948), but the lowest \textsc{ConstraintCorrect} (0.796), reflecting frequent
violations of strict output-form requirements. Task~C shows a different
bottleneck: it has the highest \textsc{ConstraintCorrect} (0.936) but the lowest
\textsc{SemanticCorrect} (0.816), because controlled administrative updates
create more opportunities for omitting, distorting, or incompletely realizing
the intended update.

\begin{figure}[!t]
  \centering
  \includegraphics[width=\columnwidth]{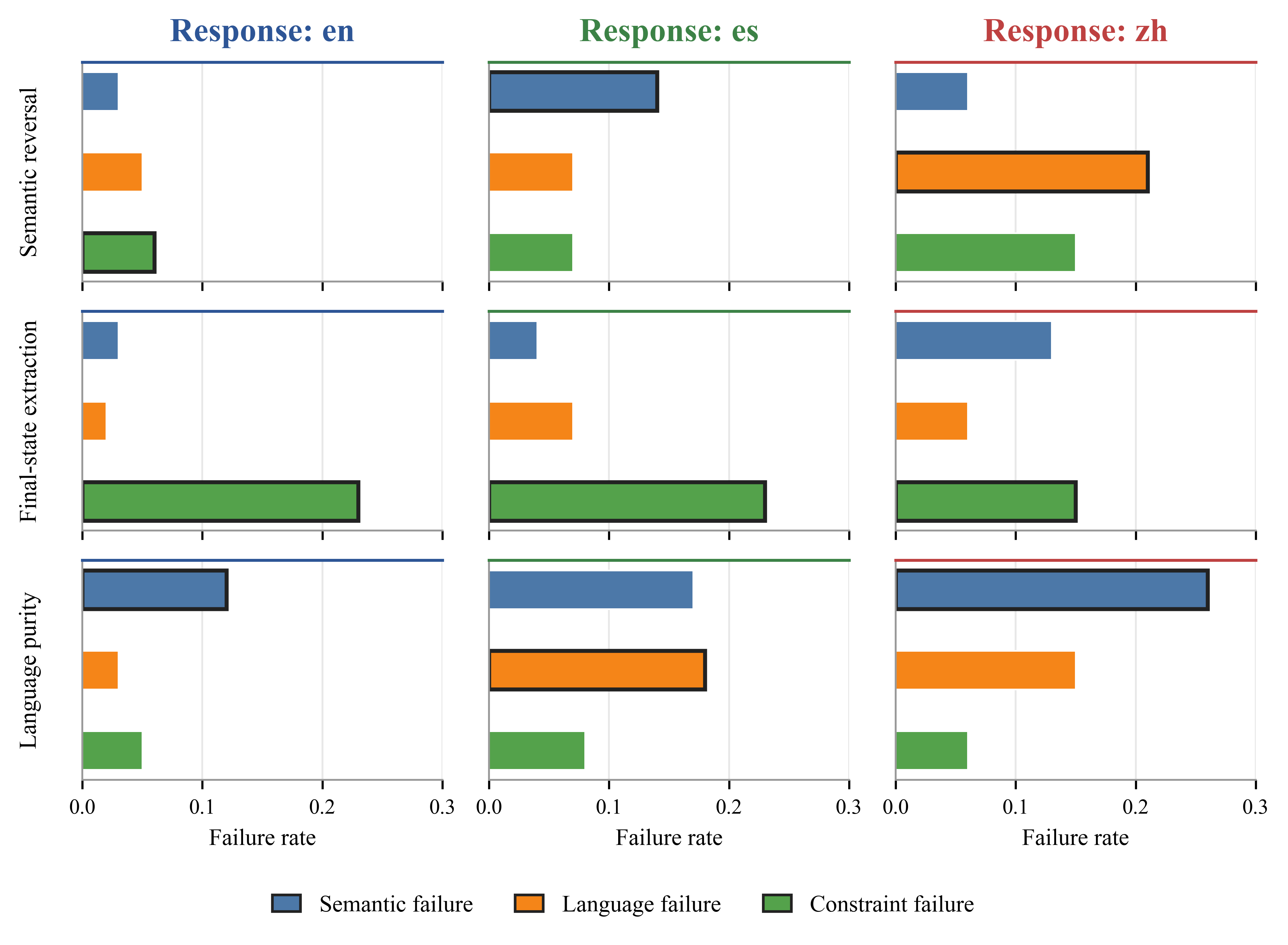}
\caption{Component failure profiles by task family and response language.
Bars show one minus each score; outlines mark the dominant failure component.}
  \label{fig:failure_profile}
\end{figure}

Two deviations from the overall response-language pattern are visible in
Figure~\ref{fig:cat_resp}. First, Task~B is the only family in which ZH-response
performance does not degrade relative to EN-response performance
(\textsc{JointSuccess}=0.746 for ZH and 0.743 for EN), consistent with its short
temporal, numeric, or named-entity answers. Second, Task~C shows a marked language-control failure on ES-response items
(Figure~\ref{fig:failure_profile}), driven by retained English structural tokens
such as \textit{Room}, \textit{Building}, and \textit{Hall}. Its overall
\textsc{JointSuccess}, however, remains lowest for ZH-response items.

Figure~\ref{fig:failure_profile} makes these bottlenecks explicit. Although the
three task families have similar \textsc{JointSuccess}, their component-level
failure profiles differ sharply. Final-state extraction is dominated by
constraint failures across response languages. Semantic reversal with ZH
responses is mainly limited by target-language realization. Language-purity
tasks show a different pattern: ZH responses are dominated by semantic update
failures, whereas ES responses combine semantic and language-control failures.

\FloatBarrier

\section{Conclusion}

MTM-Bench frames multilingual task execution as a triplet-structured problem in
which instruction, content, and response languages vary independently. This
design exposes language-role effects that are hidden when benchmarks aggregate
over language configurations.

Across 20 models and 27 language triplets, the response language is the main
source of variation, while instruction and content languages play smaller and
less consistent roles. The response-only and full-mismatch comparison suggests that
difficulty is not governed by mismatch count alone: once the requested answer
language departs from the preceding context, extra mismatches add little
explanatory power. Because this ordering varies across models, we treat it as
evidence against a monotonic mismatch-count account, not as a directional claim
about full mismatch.

Task-family results show why decomposed evaluation matters. Similar end-to-end
scores can reflect different failures, including correct meanings expressed in
the wrong language and target-language outputs that violate format requirements.
Multilingual evaluation should therefore track the role each language plays
rather than collapse distinct configurations into one aggregate score. Without
this distinction, aggregate scores may obscure failures in the specific language
role required by the task.

\section*{Limitations}

MTM-Bench covers EN, ES, and ZH across three task families, fully crossing
instruction, content, and response language roles while keeping task semantics
and scoring rules tightly controlled. The observed response-role dominance
should therefore be interpreted as a controlled finding within the EN/ES/ZH
setting rather than as a universal claim about all multilingual task execution.
EN and ES provide a high-resource same-script contrast, while ZH provides one
non-Latin-script contrast. Whether the same role structure changes with
typological distance, script, or resource level remains an open question.
Broader language coverage and additional task families would be valuable for
testing whether the role-position effect and response-axis pattern extend to
more typologically distinct or lower-resource languages.

With 30 base items per task family, MTM-Bench is better suited for controlled
role-specific diagnosis than for estimating broad domain-level multilingual
performance. The language-purity family, in particular, draws on short
administrative update scenarios; its findings are best interpreted as evidence
about localized update realization under controlled conditions, not as a
comprehensive account of update-generation tasks more generally.

Automated scoring remains imperfect for short extractions, mixed-script
responses, paraphrastic answers, aliases, and localized entity names. The
combination of deterministic checks, LLM-assisted semantic judgment, and a
stratified human audit reduces but does not eliminate residual errors in
boundary cases. The main remaining risk is semantic labeling for Families A and
C, where \textsc{SemanticCorrect} is assigned by a single LLM-assisted judge.
This judge is blinded to model identity, and its role is limited to semantic
content: \textsc{LangCorrect}, \textsc{ConstraintCorrect},
\textsc{ContaminationRatio}, and all Family B semantic labels are determined
outside the LLM judge. Although the judge model is also included among the
evaluated systems, model-family information is not provided to the judge. This
risk is further bounded by the targeted human audit and the unflagged sanity
check, which directly inspect boundary cases and routine rows. Multi-judge sensitivity analysis and expanded human annotation would further
characterize this residual risk.

MTM-Bench should therefore be viewed as a controlled diagnostic probe rather
than a comprehensive multilingual leaderboard. To support reproducibility, we
include the benchmark data, evaluation code, scored model outputs, and bootstrap
scripts in the supplementary material. The same triplet-based framework can be
extended to additional languages, domains, and task types, enabling future work
to test whether the role-specific patterns observed here hold in broader
multilingual settings.

\section*{Ethical Considerations}

MTM-Bench is constructed from synthetic administrative and evaluative scenarios
and does not contain personal, private, user-derived, or intentionally offensive
content. The benchmark is
intended for controlled evaluation of multilingual task execution rather than
for deployment in high-stakes decision-making settings. Because the benchmark
uses short multilingual prompts and automatic scoring, residual scoring errors
may occur, especially near language-boundary and paraphrase cases; we mitigate
this risk through decomposed scoring and a human audit. The human audit was
conducted internally by members of the research team to validate scoring
decisions and did not involve recruited external participants or crowdworkers.

The released artifacts are intended for research use and include synthetic benchmark instances, evaluation scripts, aggregate analyses, and scored model outputs. Closed-source models were accessed through official provider APIs or public interfaces under the applicable provider terms, and open-weight models were used through released checkpoints or hosted inference endpoints under the licenses stated by their creators. We do not redistribute model weights, provider documentation, or provider system artifacts; included model outputs are provided only as benchmark evaluation records.

\bibliography{ref}

\clearpage
\appendix
\section{Generation Settings and Benchmark Inventory}
\label{app:generation}

All model outputs are generated in a zero-shot setting with no system prompt and
no few-shot demonstrations~\citep{brown2020language,wei2022finetuned}. For each benchmark instance, we use a fixed prompt
template that concatenates the localized instruction, source content, and
question:
\begin{quote}
\small
\texttt{instruction\_text}

\texttt{Content: content\_text}

\texttt{Question: question}
\end{quote}
We request one response per model. In our generation scripts, we set temperature
to 0 when the provider supports it and leave top-$p$ at the provider default~\citep{holtzman2020curious}
because decoding controls are not exposed uniformly across APIs. We do not
enable optional reasoning or thinking modes unless they are part of the model's
default public interface.

The benchmark contains 90 base items in total: 30 semantic-reversal items,
30 final-state extraction items, and 30 language-purity items. Each base item is
instantiated under all 27 language triplets, yielding 810 instances per family
and 2{,}430 instances per model. Semantic-reversal and language-purity items use
LLM-assisted semantic judging, while final-state extraction uses a deterministic
final-value checker.

\section{Additional Model-Level Results}
\label{app:model_family}

Table~\ref{tab:model_appendix} reports model-level task-family performance and
token usage. The ranking across task families is not monotone: models that
perform well on semantic reversal do not always lead on final-state extraction
or language purity, reflecting the distinct capability profiles exposed by each
family. Some models generate substantially longer outputs, but higher output
length does not uniformly correspond to higher \textsc{JointSuccess}.

\begin{figure}[!h]
  \centering
  \includegraphics[width=\columnwidth]{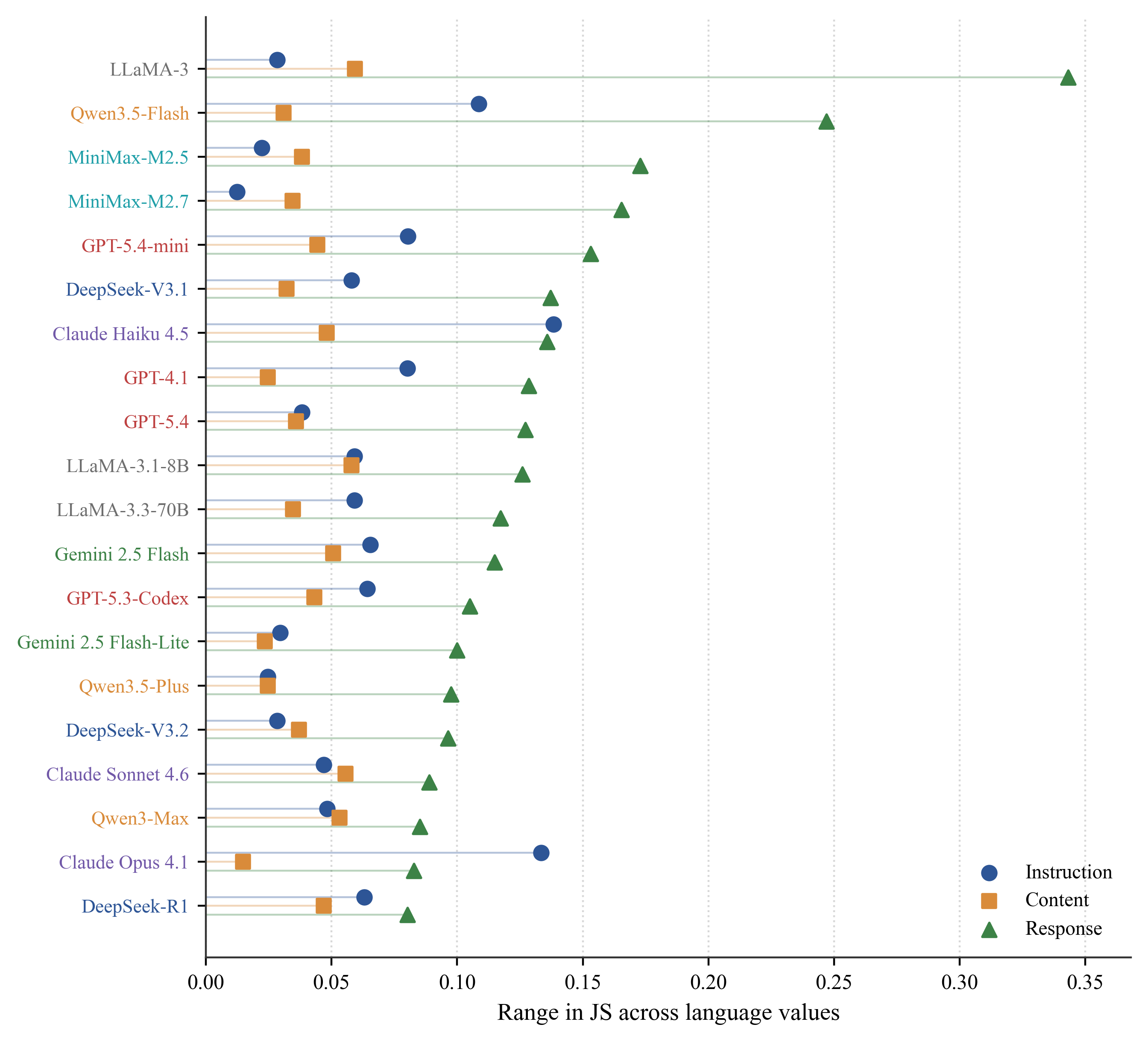}
  \caption{Model-level sensitivity to the three language roles. Each point
  shows the range in \textsc{JointSuccess} across English, Spanish, and
  Chinese for one language slot, averaging over the other two slots.}
  \label{fig:model_role_sensitivity}
\end{figure}

\begin{figure}[!h]
  \centering
  \includegraphics[width=\columnwidth]{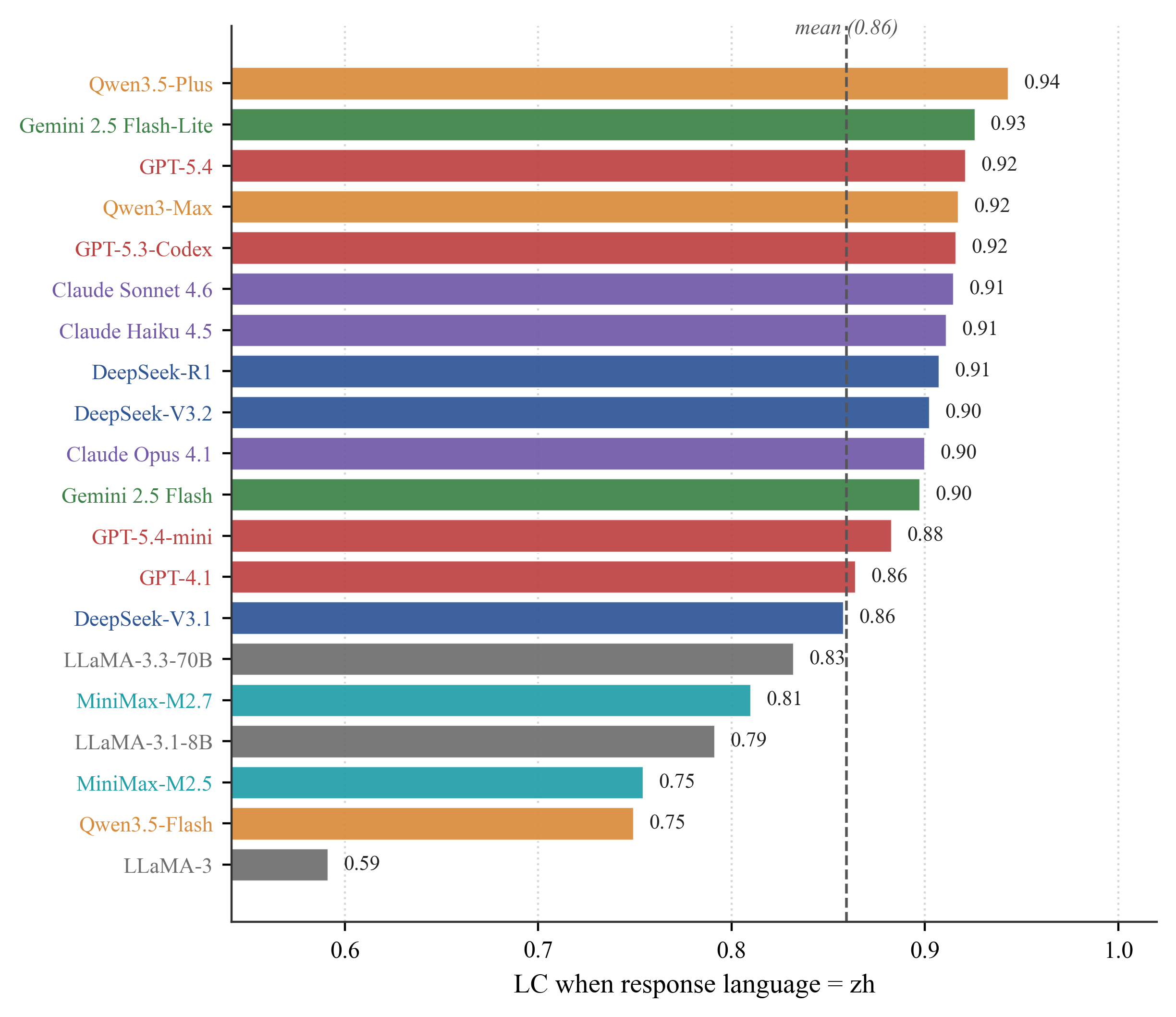}
  \caption{\textsc{LangCorrect} on Chinese-response instances, sorted by
  model.}
  \label{fig:zh_lang_rank}
\end{figure}

Figure~\ref{fig:model_role_sensitivity} shows that the response role produces
the largest model-level sensitivity for many systems. Figure~\ref{fig:zh_lang_rank}
isolates \textsc{LangCorrect} on Chinese-response instances, highlighting which
models are most robust under the most difficult response-language setting.

\begin{table*}[!t]
\centering
\footnotesize
\setlength{\tabcolsep}{3.2pt}
\renewcommand{\arraystretch}{1.03}
\begin{tabular}{@{}lrrr@{\hspace{1.4em}}lrrr@{}}
\toprule
\multicolumn{4}{c}{\textbf{Task-family JS}} &
\multicolumn{4}{c}{\textbf{Token usage}} \\
\cmidrule(r){1-4}\cmidrule(l){5-8}
\textbf{Model} & \textbf{A} & \textbf{B} & \textbf{C}
& \textbf{Model} & \textbf{In} & \textbf{Out} & \textbf{JS} \\
\midrule
Qwen3-Max             & .891 & \best{.801} & \best{.826}
& Qwen3-Max             &  77.8 &  20.4 & \best{.840} \\
DeepSeek-R1           & .894 & \best{.804} & .769
& DeepSeek-R1           &  77.8 & 344.4 & .822 \\
Qwen3.5-Plus          & \best{.927} & .770 & .763
& Qwen3.5-Plus          &  80.7 &  18.7 & .820 \\
GPT-5.4-mini          & .878 & .754 & .814
& GPT-5.4-mini          &  74.4 &  51.2 & .815 \\
GPT-5.4               & \best{.915} & .709 & .773
& GPT-5.4               &  74.4 &  46.0 & .799 \\
DeepSeek-V3.2         & .863 & .756 & .769
& DeepSeek-V3.2         &  74.7 &  18.2 & .796 \\
DeepSeek-V3.1         & .830 & .778 & .749
& DeepSeek-V3.1         &  72.8 &  18.4 & .786 \\
GPT-5.3-Codex         & \best{.928} & \worst{.631} & .784
& GPT-5.3-Codex         &  74.4 &  71.8 & .781 \\
Claude Sonnet 4.6     & .804 & .722 & .776
& Claude Sonnet 4.6     &  97.2 &  34.3 & .767 \\
Gemini 2.5 Flash-Lite & .848 & .759 & .694
& Gemini 2.5 Flash-Lite &  72.1 &  19.1 & .767 \\
Qwen3.5-Flash         & .775 & .753 & .765
& Qwen3.5-Flash         &  80.7 &  19.5 & .765 \\
GPT-4.1               & .848 & .683 & .742
& GPT-4.1               &  75.4 &  18.6 & .758 \\
Claude Opus 4.1       & .730 & .780 & .712
& Claude Opus 4.1       &  97.2 &  33.7 & .741 \\
LLaMA-3.3-70B         & .815 & .700 & .704
& LLaMA-3.3-70B         & 109.0 &  20.7 & .740 \\
Gemini 2.5 Flash      & .733 & \best{.779} & .669
& Gemini 2.5 Flash      &  72.1 &  19.7 & .727 \\
MiniMax-M2.7          & .746 & \best{.791} & .619
& MiniMax-M2.7          &  91.8 & 256.4 & .719 \\
MiniMax-M2.5          & .775 & .760 & \worst{.507}
& MiniMax-M2.5          &  99.9 & 297.5 & .681 \\
Claude Haiku 4.5      & \worst{.527} & .747 & .759
& Claude Haiku 4.5      &  97.2 &  37.7 & .678 \\
LLaMA-3               & \worst{.538} & \worst{.577} & .604
& LLaMA-3               &  84.0 &  27.2 & .573 \\
LLaMA-3.1-8B          & \worst{.464} & \worst{.500} & .646
& LLaMA-3.1-8B          &  84.0 &  18.9 & .537 \\
\midrule
\textit{Mean}         & .786 & .728 & .722
& \textit{Mean}         &  83.4 &  69.6 & .745 \\
\bottomrule
\end{tabular}
\caption{Additional model-level results. Left: \textsc{JointSuccess} by task
family, where A~=~semantic reversal, B~=~final-state extraction, and
C~=~language purity. Right: mean input and output tokens per instance. The
bottom row reports unweighted averages over models.}
\label{tab:model_appendix}
\end{table*}

\subsection{Model-level Heterogeneity in Response-only and Full-mismatch Triplets}
\label{app:mismatch_heterogeneity}

The aggregate response-only and full-mismatch comparison should not be
interpreted as a universal model-level ordering. Across the 20 evaluated models,
12 perform worse on response-only triplets than on full-mismatch triplets, 7
show the opposite ordering, and 1 is tied or nearly tied. The aggregate
full-minus-response-only difference is 0.011 under clustered bootstrap
resampling over base items. We therefore interpret this result as evidence that
mismatch count is not a monotonic predictor of difficulty, rather than as a
directional claim that full mismatch is intrinsically easier.

\section{Additional Triplet-Level Visualizations}
\label{app:triplet_visuals}

\begin{figure*}[!h]
  \centering
  \includegraphics[width=0.94\textwidth]{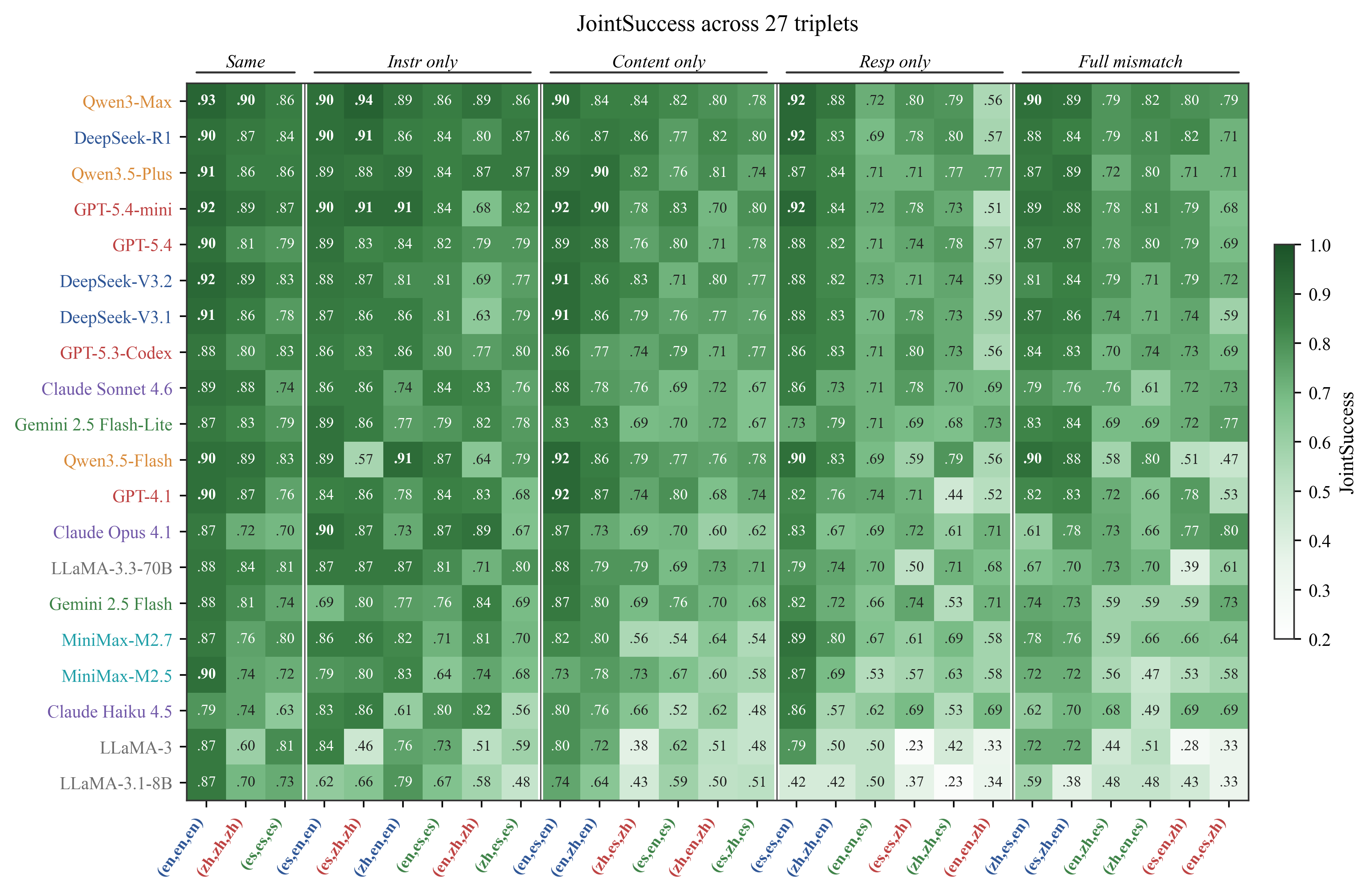}
  \caption{\textsc{JointSuccess} across the 27 triplets, grouped by mismatch
  class and ordered within each class by mean \textsc{JointSuccess}. Darker
  green indicates higher joint success; tick-label colors indicate response
  language.}
  \label{fig:heatmap}
\end{figure*}

Figure~\ref{fig:heatmap} presents \textsc{JointSuccess} for all 27 language
triplets across 20 models. Triplets are grouped by mismatch class and ordered
by mean \textsc{JointSuccess} within each group, making it straightforward to
compare the within-class spread against the between-class gap. The within-class
ordering shows that even among triplets sharing the same mismatch type, the
choice of specific languages still introduces visible variation, especially in
the response-only and full-mismatch groups.

\section{Scoring Edge Cases and Task Rubrics}
\label{app:scoring_details}

\subsection{LangCorrect Edge Cases}
\label{app:edge}

Table~\ref{tab:edge_cases} summarizes representative boundary decisions used
when applying the target-language criterion.

\begin{table}[!h]
\centering
\footnotesize
\setlength{\tabcolsep}{3.0pt}
\renewcommand{\arraystretch}{1.08}
\begin{tabularx}{\columnwidth}{@{}
  >{\raggedright\arraybackslash}p{0.33\columnwidth}
  c
  c
  >{\RaggedRight\arraybackslash}X
@{}}
\toprule
\textbf{Output} & \textbf{Target} & \textbf{LC} & \textbf{Reason} \\
\midrule
\texttt{327}
  & zh & 1 & Pure numeral. \\
\texttt{Rivera}
  & zh & 1 & Allowlisted proper name. \\
\texttt{Professor Rivera}
  & zh & 0 & English title not localized. \\
\texttt{Professor Rivera}
  & es & 0 & English title spelling; Spanish uses profesor/profesora. \\
\texttt{Profesor Rivera}
  & es & 1 & Localized title plus proper name. \\
\texttt{Friday at 4 PM}
  & zh & 0 & English phrase in a Chinese response. \\
\texttt{Room 327}
  & zh & 0 & English structural token. \\
\zh{327室}
  & zh & 1 & Localized room expression. \\
\texttt{Hall 2}
  & es & 0 & English structural token. \\
\texttt{el sal\'on 2}
  & es & 1 & Localized Spanish expression. \\
\bottomrule
\end{tabularx}
\caption{Representative \textsc{LangCorrect} decisions for short extraction
outputs.}
\label{tab:edge_cases}
\end{table}

\subsection{Task-Specific Semantic Rubrics}
\label{app:rubric}

The three task families apply different semantic correctness criteria,
reflecting the distinct challenges each one targets~\citep{vanhee2018irony,jiang2024followbench,marchisio2024languageconfusion}.

\paragraph{Family A: Semantic reversal.}
An output is scored as correct if it recovers the author's actual stance
rather than the surface framing. Wording may differ from the reference
answer, but polarity and the target of evaluation must be preserved.

\paragraph{Family B: Final-state extraction.}
An output is scored as correct if it returns the final valid value after
updates, reschedules, reassignments, or revisions. Earlier mentioned values
receive no credit, even if they were valid at an intermediate stage.

\paragraph{Family C: Language purity and update realization.}
An output is scored as correct if it captures the main update or status
change without contradiction. Minor paraphrastic variation is allowed, and
omission of secondary details is acceptable so long as the core update is
preserved.

\section{LLM-Assisted Semantic Judging Details}
\label{app:llm_judge}

We use LLM-assisted semantic judging only for task families whose semantic
correctness cannot be fully captured by deterministic string or anchor
matching~\citep{kocmi-federmann-2023-large,liu2023geval,zheng2023judging}. Families A and C are scored with this procedure, while Family~B is
scored by a deterministic final-state checker. This separation avoids using a
generative judge when the correct value can be verified by structured rules,
such as dates, times, room numbers, contacts, or localized numeric variants.

For Family~A, the judge determines whether the output recovers the author's
actual stance rather than the surface framing, focusing on whether both the
polarity and the evaluative target match the reference. For Family~C, the
judge determines whether the output captures the main update or status change
without contradiction; minor paraphrases and omissions of secondary details
are permitted when they do not alter the core update.

The judge scores only semantic content. Target-language adherence,
code-switching, formatting, sentence count, and output-length constraints are
handled separately by \textsc{LangCorrect}, \textsc{ContaminationRatio}, and
\textsc{ConstraintCorrect}. This separation matters because a response may be
semantically correct while retaining an English structural shell, or fluent
in the target language while missing the required update~\citep{rahmati2025cococola}.

The LLM judge returns a binary label,
\[
\textsc{SemanticCorrect}(\hat{y},\, y^{\star}) \in \{0,1\}.
\]
The prompt includes the task family, the reference answer or rubric, and the
model output. The judge does not receive the identity of the model that produced
the output, preventing model-family information from influencing the semantic
label. Although the judge model is also included among the evaluated systems, the
semantic judge is blinded to model identity and is used only for
\textsc{SemanticCorrect} in Families A and C; \textsc{LangCorrect},
\textsc{ConstraintCorrect}, \textsc{ContaminationRatio}, and all Family B
semantic labels are determined outside the LLM judge. We use Claude Haiku 4.5 as the semantic judge at temperature~0.

We use the following judge prompt template for Families A and C:

\begin{quote}
\small
You are a factual evaluator for a multilingual benchmark. Judge only
semantic correctness. Ignore language purity, code-switching, and formatting
constraints.

For Family~A, the gold answer captures the author's real stance, not the
surface wording. Do not reward literal praise if the passage is actually
critical. The output is correct only if the polarity and the target of
criticism are both correct.

For Family~C, mark the output correct if the core update is captured
accurately and there is no contradiction. Minor paraphrase, mixed-language
wording, or non-target-language shells do not make it semantically wrong.
Score only semantic content.

Reply with only the digit 1 or 0.
\end{quote}

This design keeps semantic judging narrow and decoupled from the stricter
language-purity and constraint checks. The LLM-assisted judge therefore
serves as a targeted semantic adjudicator rather than as a replacement for
the rule-based scoring protocol.

\section{Human Scoring Audit}
\label{app:human}

We conduct a human audit to assess the reliability of the scoring pipeline and
to characterize recurring boundary cases. The audit includes two complementary sets: a targeted audit set containing
1,022 cases marked during scoring development and validation for closer review,
and an additional unflagged sanity check of 300 outputs sampled from routine
rows without review flags.

For each audited output, annotators inspect the benchmark item, model response,
and automatic component labels. The audit records agreements and human overrides
for the binary component labels: \textsc{SemanticCorrect},
\textsc{LangCorrect}, and \textsc{ConstraintCorrect}. Rows without human
overrides are counted as agreements, and overrides are adjudicated using the
edge-case policy in Appendix~\ref{app:edge}~\citep{artstein2008inter}.

Table~\ref{tab:human_audit_summary} summarizes the size of each audit subset
and the corresponding component-level and \textsc{JointSuccess}-level override
rates.

\begin{table}[t]
\centering
\footnotesize
\setlength{\tabcolsep}{2.6pt}
\renewcommand{\arraystretch}{1.04}
\begin{tabular}{@{}lrrrr@{}}
\toprule
\textbf{Subset} & $n$ & \textbf{Comp.} & \textbf{Rate} & \textbf{JS} \\
\midrule
Targeted audit set & 1,022 & 186 & 18.2\% & 15.4\% \\
Unflagged check & 300 & 22 & 7.3\% & 7.3\% \\
\bottomrule
\end{tabular}
\caption{Human audit summary. Comp. reports the number of audited rows with at
least one override to \textsc{SemanticCorrect}, \textsc{LangCorrect}, or
\textsc{ConstraintCorrect}. JS reports the percentage of rows whose final
\textsc{JointSuccess} label changes.}
\label{tab:human_audit_summary}
\end{table}

Agreement rates vary across response languages, with the largest differences
appearing in \textsc{SemanticCorrect} and \textsc{ConstraintCorrect}, while
\textsc{LangCorrect} remains stable across all three languages
(Table~\ref{tab:human_audit_by_language}).

\begin{table}[t]
\centering
\small
\setlength{\tabcolsep}{4pt}
\begin{tabular}{@{}lrrrr@{}}
\toprule
\textbf{Resp. lang.} & \textbf{Overall} & \textbf{SC} & \textbf{LC} & \textbf{CC} \\
\midrule
EN & 86.9\% & 96.1\% & 99.4\% & 90.8\% \\
ES & 80.4\% & 89.6\% & 98.8\% & 91.4\% \\
ZH & 78.8\% & 89.3\% & 98.0\% & 89.0\% \\
\bottomrule
\end{tabular}
\caption{Human-audit agreement rates by response language and scoring
component, computed on the targeted audit set. Overall agreement requires
agreement on all three binary component labels.}
\label{tab:human_audit_by_language}
\end{table}

The unflagged sanity check yields a lower override rate than the targeted audit
set, suggesting that most routine cases are handled consistently by the
automatic pipeline. Overrides concentrate in a small number of boundary
conditions, especially strict \texttt{only\_output} constraints, paraphrastic
semantic matches, and target-language realization near the allowlist boundary.
These cases are precisely where decomposed scoring is useful: semantic,
language, and constraint errors can be corrected independently rather than
collapsed into a single undifferentiated label.

The \textsc{JointSuccess}-level override rate is lower than the component-level
override rate because some component corrections do not change the final joint
decision. For example, correcting a language label does not affect
\textsc{JointSuccess} when the same response already fails an explicit format
constraint. Overall, the audit supports the decomposed scoring policy while
motivating caution for small differences between closely ranked models.

\begin{figure*}[!t]
  \centering
  \includegraphics[width=0.92\textwidth]{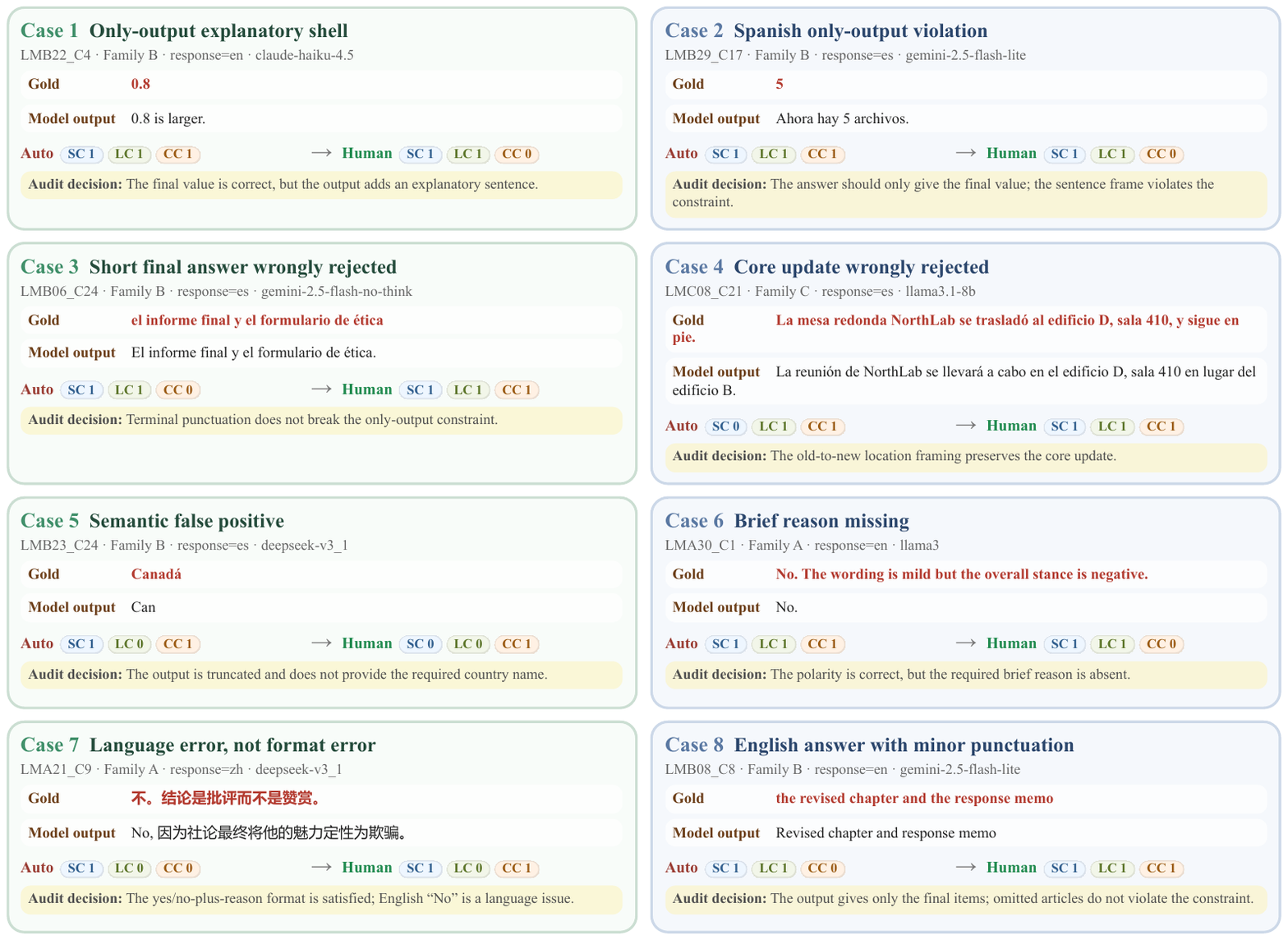}
  \caption{Representative human-audit cases comparing automatic labels with
  adjudicated human labels for common scoring boundary cases.}
  \label{fig:human_case}
\end{figure*}

Figure~\ref{fig:human_case} presents representative cases from the audit,
showing the automatic labels alongside the adjudicated human labels. Cases are
selected to illustrate common override sources: output-form violations,
punctuation-level constraint boundaries, paraphrastic semantic matches, and
old-to-new transition framing that preserves the core update despite surface
differences from the gold reference.

\section{Statistical Uncertainty Analysis}
\label{app:stat_uncertainty}

To quantify uncertainty in aggregate benchmark estimates, we use clustered
bootstrap resampling over base items~\citep{efron1994introduction,
koehn2004statistical,dror2018hitchhiker}. Each base item appears in all 27 language triplet
configurations for each model, so resampling individual rows would treat
triplet variants derived from the same underlying item as independent
observations and would underestimate item-level uncertainty. In each
bootstrap replicate, we sample base items with replacement and retain all
associated triplet instances and model outputs. We then recompute aggregate
quantities, including model-level metric means, mismatch-class means,
task-family means, and marginal language-axis ranges. Reported 95\% intervals
are percentile intervals over bootstrap replicates.

The marginal range in \textsc{JointSuccess} is largest for the
response-language axis. The response-axis range is 0.113, with a 95\%
clustered bootstrap interval of $[0.083, 0.156]$, compared with
$0.029\ [0.012, 0.049]$ for the content axis and
$0.018\ [0.006, 0.033]$ for the instruction axis
(Table~\ref{tab:bootstrap_axis_ranges}). The separation among these intervals
supports the interpretation that response-language dominance is not an artifact
of item-level sampling variance and is consistent with the main-text finding
that response-language control is the dominant source of variation in this
benchmark.

\begin{table}[t]
\centering
\small
\setlength{\tabcolsep}{4pt}
\renewcommand{\arraystretch}{1.04}
\begin{tabular}{@{}lcc@{}}
\toprule
\textbf{Axis} & \textbf{Range} & \textbf{95\% CI} \\
\midrule
Instruction & 0.018 & $[0.006, 0.033]$ \\
Content     & 0.029 & $[0.012, 0.049]$ \\
Response    & 0.113 & $[0.083, 0.156]$ \\
\bottomrule
\end{tabular}
\caption{Clustered bootstrap intervals for language-axis ranges in
\textsc{JointSuccess}.}
\label{tab:bootstrap_axis_ranges}
\end{table}

\paragraph{Mismatch-class comparison.}
We use the same base-item clustered bootstrap to quantify uncertainty in the
comparison between response-only and full-mismatch triplets. For this analysis,
the statistic is the \textsc{JointSuccess} difference
\[
\Delta_{\mathrm{full-resp}}
=
\textsc{JS}_{\mathrm{full\ mismatch}}
-
\textsc{JS}_{\mathrm{resp\ only}}.
\]
The observed response-only mean is 0.694 and the observed full-mismatch mean is
0.705. The resulting full-minus-response-only difference is 0.011, with a
95\% clustered bootstrap interval of [0.001, 0.022]. Although the interval lies
slightly above zero, the difference is small and runs counter to the direction
expected under a monotonic mismatch-count account: full-mismatch triplets do not
score below response-only triplets. Because the response-only versus
full-mismatch ordering is not consistent across models
(Appx.~\ref{app:mismatch_heterogeneity}), we interpret this comparison as
evidence that mismatch count is not a monotonic predictor of difficulty, rather
than as a directional claim that full mismatch is intrinsically easier.

\section{Regression Analysis with Model and Item Controls}
\label{app:regression_analysis}

We conduct an additional regression analysis to verify that the
response-language effect is not driven solely by differences among models or
by variation in base-item difficulty. We fit a fixed-effect linear probability
model for \textsc{JointSuccess}:
\[
\begin{aligned}
Y_{mijt}
  &= \beta_0
   + \alpha_{L_{\mathrm{instr}}}
   + \gamma_{L_{\mathrm{content}}}
   + \delta_{L_{\mathrm{resp}}} \\
  &\quad + \mu_m
   + \nu_i
   + \varepsilon_{mijt},
\end{aligned}
\]
where $Y_{mijt}$ denotes \textsc{JointSuccess} for model $m$, base item
$i$, and language triplet $t$. The terms $\mu_m$ and $\nu_i$ are model and
base-item fixed effects, respectively. Standard errors are clustered by base
item~\citep{cameron2015practitioners}, and English serves as the reference
level for each language slot.

Table~\ref{tab:regression_axis_effects} reports the estimated language-axis
coefficients. Response-language coefficients are substantially larger in
magnitude than those for the instruction or content language: requiring a
Spanish or Chinese response rather than an English one is associated with
decreases of 0.108 and 0.113 in \textsc{JointSuccess}, respectively.
Instruction-language coefficients are close to zero, while content-language
effects are smaller and not directionally consistent across ES and ZH. This
pattern supports the main finding that response-language variation is the
dominant source of role-specific degradation after controlling for model
identity and base-item difficulty.

\begin{table*}[!h]
\centering
\small
\setlength{\tabcolsep}{8pt}
\renewcommand{\arraystretch}{1.05}
\begin{tabular*}{0.78\textwidth}{@{\extracolsep{\fill}}lccc@{}}
\toprule
\textbf{Predictor}
  & \textbf{Coef.}
  & \textbf{95\% CI lower}
  & \textbf{95\% CI upper} \\
\midrule
Instruction: Spanish vs.\ English &  0.011 & $-$0.005 &  0.026 \\
Instruction: Chinese vs.\ English & $-$0.007 & $-$0.026 &  0.012 \\
Content: Spanish vs.\ English     &  0.029 &  0.011   &  0.047 \\
Content: Chinese vs.\ English     &  0.021 & $-$0.003 &  0.044 \\
Response: Spanish vs.\ English    & $-$0.108 & $-$0.147 & $-$0.070 \\
Response: Chinese vs.\ English    & $-$0.113 & $-$0.153 & $-$0.073 \\
\bottomrule
\end{tabular*}
\caption{Fixed-effect linear probability model for \textsc{JointSuccess}.
The model controls for model identity and base item; standard errors are
clustered by base item. English is the reference level.}
\label{tab:regression_axis_effects}
\end{table*}

\section{Benchmark Items and Representative Output Cases}
\label{app:items}

\begin{figure*}[!h]
  \centering
  \includegraphics[page=1,width=0.96\textwidth]{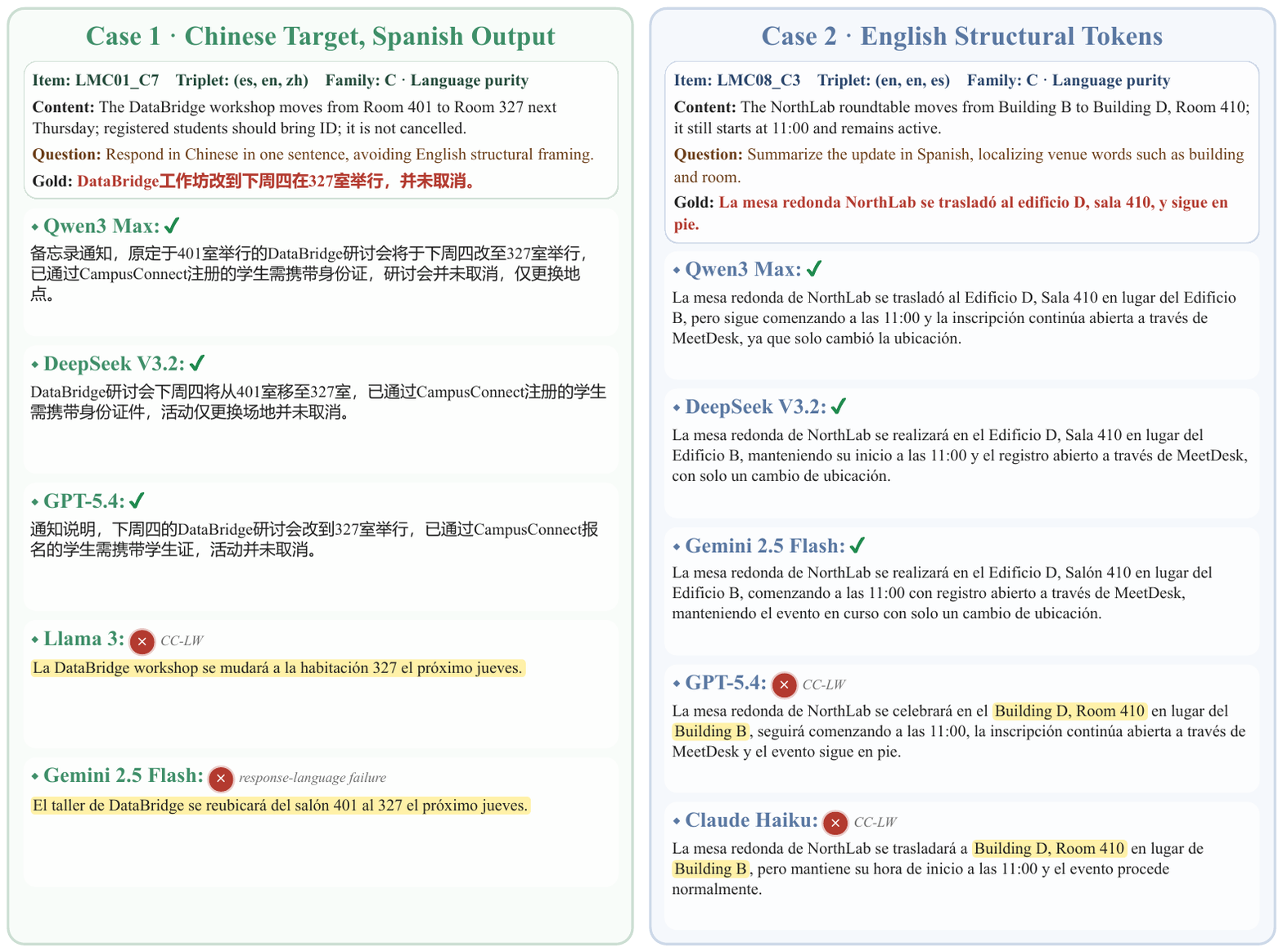}
  \caption{Representative output cases for language-purity tasks. Case~1
  shows a Chinese-response item in which some models identify the correct
  update but respond in Spanish. Case~2 shows Spanish responses that preserve
  the correct update while retaining English structural tokens.}
  \label{fig:cases12}
\end{figure*}

\begin{figure*}[p]
  \centering
  \includegraphics[page=2,width=0.96\textwidth]{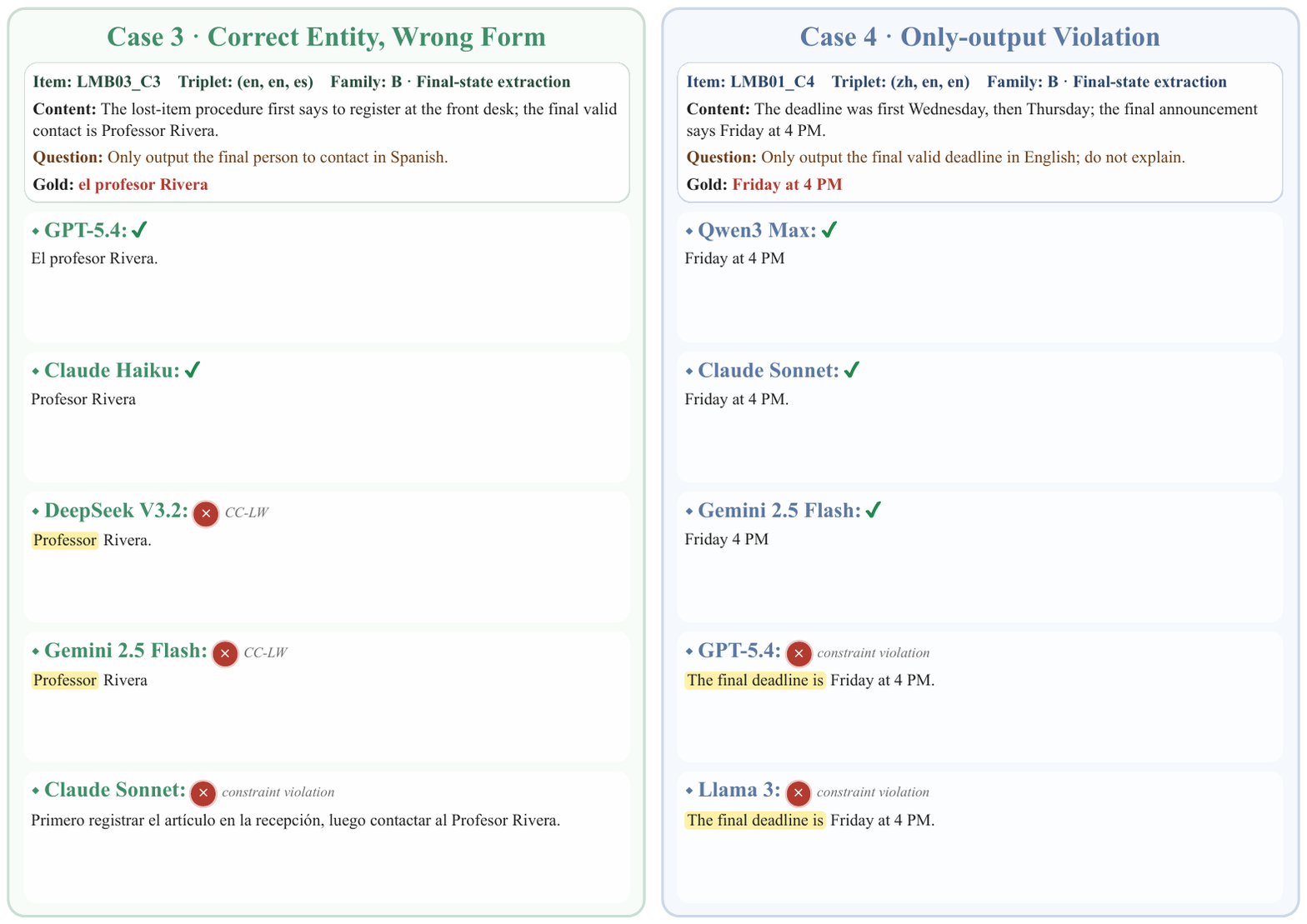}
  \caption{Representative output cases for final-state extraction. Case~3
  shows correct identification of the final contact together with failures in
  target-language realization or output form. Case~4 shows
  \texttt{only\_output} violations surrounding a semantically correct final
  deadline.}
  \label{fig:cases34}
\end{figure*}

\begin{figure*}[p]
  \centering
  \includegraphics[page=3,width=0.96\textwidth]{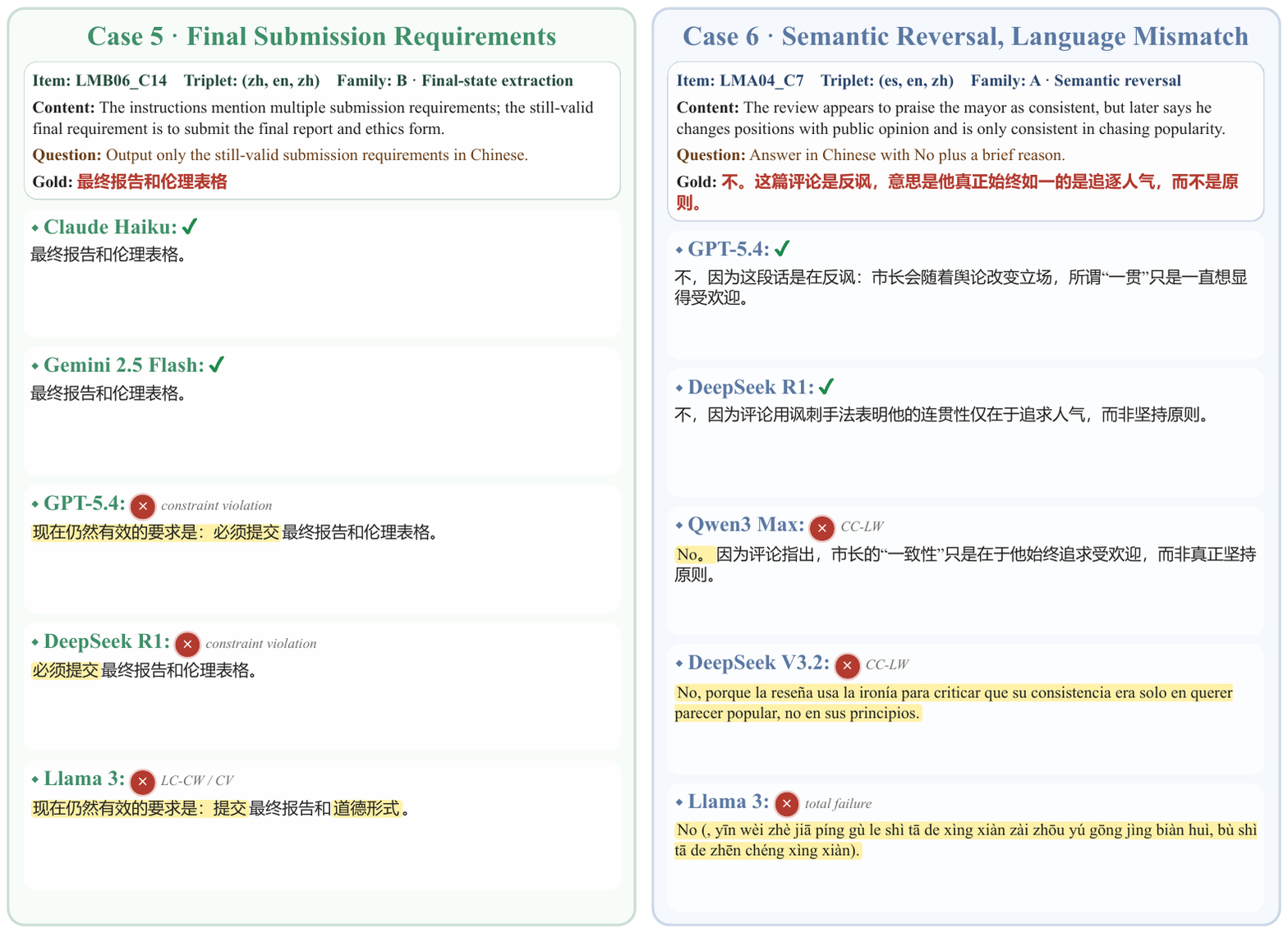}
  \caption{Representative output cases for final-state extraction and
  semantic reversal. Case~5 shows submission-requirement outputs where some
  models identify the correct final requirement but violate the requested
  output form. Case~6 shows semantic-reversal outputs with failures in the
  required response language.}
  \label{fig:cases56}
\end{figure*}

\begin{figure*}[p]
  \centering
  \includegraphics[page=4,width=0.96\textwidth]{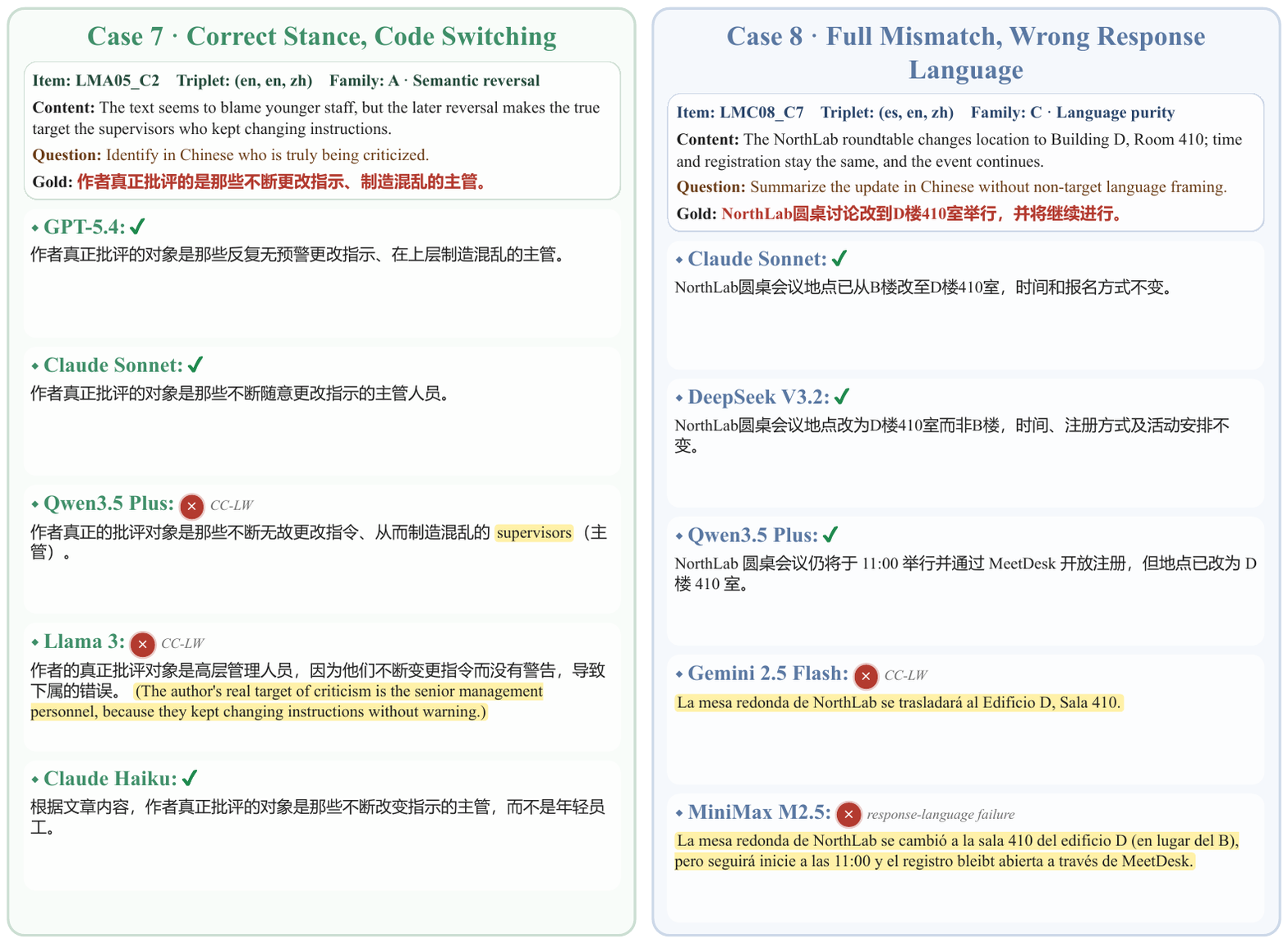}
  \caption{Representative output cases for semantic reversal and full
  mismatch. Case~7 illustrates code-switching in a Chinese semantic-reversal
  item, while Case~8 shows full-mismatch examples where the update is
  semantically close but produced in the wrong response language.}
  \label{fig:cases78}
\end{figure*}

Table~\ref{tab:item_examples} shows one representative item from each task
family, illustrating how the instruction language, content language, and
required response language are specified independently within a single
instance. Figures~\ref{fig:cases12}--\ref{fig:cases78} then present
representative model outputs from these and analogous items, showing how
semantic correctness, target-language adherence, explicit constraint
satisfaction, and code-switching can come apart under multilingual task
execution. Taken together, the items and output cases ground the aggregate
metric results in concrete examples of where and how models succeed or fail.

\begin{table*}[!h]
\centering
\small
\setlength{\tabcolsep}{5pt}
\renewcommand{\arraystretch}{1.12}
\begin{tabular}{@{}
  p{0.8cm}
  >{\raggedright\arraybackslash}p{5.1cm}
  >{\raggedright\arraybackslash}p{4.1cm}
  >{\raggedright\arraybackslash}p{4.0cm}
@{}}
\toprule
\textbf{Fam.}
  & \textbf{Content snippet}
  & \textbf{Instruction example}
  & \textbf{Gold answer} \\
\midrule
A
  & ``At first, the essay seems to praise the professor \ldots\ what looked
    like authority was often only fear in disguise.''
  & ``Answer in Spanish with yes or no plus one brief reason.''
  & ``No. El autor critica al profesor\ldots''
    \newline {\emph{`No. The author criticizes the professor\ldots'}} \\
\addlinespace[0.3em]
B
  & ``The seminar originally scheduled for Tuesday in Room 214 will now be
    held online one hour later.''
  & ``Only output the final valid schedule in Chinese.''
  & \zh{``周二延后一小时，线上举行''}
    \newline {\emph{`Tuesday, one hour later, held online.'}} \\
\addlinespace[0.3em]
C
  & ``The workshop is moved from Room 401 to Room 327 and has not been
    cancelled.''
  & ``Respond in Chinese in one sentence. Proper nouns may remain, but
    avoid English structural framing.''
  & \zh{``研讨会改到327室举行，并未取消。''}
    \newline {\emph{`The workshop is moved to Room 327 and is not cancelled.'}} \\
\bottomrule
\end{tabular}
\caption{Representative benchmark items. English glosses of non-English gold answers are shown in italics.}
\label{tab:item_examples}
\end{table*}

% \begin{table*}[!t]
% \centering
% \small
% \setlength{\tabcolsep}{5pt}
% \renewcommand{\arraystretch}{1.10}
% \begin{tabular}{@{}
%   p{0.8cm}
%   >{\raggedright\arraybackslash}p{5.2cm}
%   >{\raggedright\arraybackslash}p{4.2cm}
%   >{\raggedright\arraybackslash}p{3.5cm}
% @{}}
% \toprule
% \textbf{Fam.}
%   & \textbf{Content snippet}
%   & \textbf{Instruction example}
%   & \textbf{Gold answer} \\
% \midrule
% A
%   & ``At first, the essay seems to praise the professor \ldots\ what looked
%     like authority was often only fear in disguise.''
%   & ``Answer in Spanish with yes or no plus one brief reason.''
%   & ``No. El autor critica al profesor\ldots'' \\
% \addlinespace[0.3em]
% B
%   & ``The seminar originally scheduled for Tuesday in Room 214 will now be
%     held online one hour later.''
%   & ``Only output the final valid schedule in Chinese.''
%   & \zh{``周二延后一小时，线上举行''} \\
% \addlinespace[0.3em]
% C
%   & ``The workshop is moved from Room 401 to Room 327 and has not been
%     cancelled.''
%   & ``Respond in Chinese in one sentence. Proper nouns may remain, but
%     avoid English structural framing.''
%   & \zh{``研讨会改到327室举行，并未取消。''} \\
% \bottomrule
% \end{tabular}
% \caption{Representative benchmark items.}
% \label{tab:item_examples}
% \end{table*}

\end{document}